\documentclass[preprint,12pt,authoryear]{elsarticle}

\usepackage[utf8]{inputenc}
\usepackage{enumitem}
\usepackage{amsmath}
\usepackage{amssymb}
\usepackage{hyperref}
\usepackage{amsfonts}
\usepackage{array}    
\usepackage{booktabs} 
\usepackage{caption}
\usepackage{color}
\usepackage{colortbl}
\usepackage[x11names, rgb]{xcolor}
\usepackage[table,xcdraw]{xcolor}
\usepackage{adjustbox}
\usepackage{tabularx}
\usepackage{graphicx}
\graphicspath{{figs/}}
\usepackage{cuted}  
\usepackage{diagbox}    
\usepackage{graphicx}
\usepackage{multicol} 
\usepackage{multirow} 
\usepackage{tabularx}
\usepackage{float}
\usepackage{pifont} 
\usepackage{caption}
\usepackage{subcaption}
\usepackage{float}
\usepackage{booktabs}
\graphicspath{{figs/}}
\usepackage{xcolor}
\usepackage{enumitem}

\journal{International Journal of Applied Earth Observation and Geoinformation}

\journal{Journal of \LaTeX\ Templates}

\biboptions{authoryear}

\begin{document}

\begin{frontmatter}

\title{Cross-Domain Transfer with Self-Supervised Spectral-Spatial Modeling for Hyperspectral Image Classification}

\tnotetext[mytitlenote]{This work was supported in part by the Supporting Project of the STS Program of the Chinese Academy of Sciences in Fujian Province under Grant 2024T3008, and in part by the Basic Scientific Research Project of the Education Department of Liaoning Province (Key Project of Independent Topic Selection, 2024) under Grant 524053218.}

\author[1,2]{Jianshu Chao}
\author[3,4]{Tianhua Lv}
\author[3,4]{Qiqiong Ma}
\author[3]{Yunfei Qiu}
\author[4]{Li Fang}
\author[1,2]{Huifang Shen\corref{cor1}}
 
\ead{shenhf@fjirsm.ac.cn} 
\author[4]{Wei Yao}

\address[1]{Quanzhou Institute of Equipment Manufacturing, Haixi Institutes, Chinese Academy of Sciences, Quanzhou, China}
\address[2]{Fujian Institute of Research on the Structure of Matter, Chinese Academy of Sciences, Fuzhou, China}
\address[3]{Liaoning Technical University, Huludao, Liaoning, China}
\address[4]{State Key Laboratory of Regional and Urban Ecology, Institute of Urban Environment, Chinese Academy of Sciences, Xiamen, China}

\cortext[cor1]{Corresponding author}

\begin{abstract}
Self-supervised learning has demonstrated considerable potential in hyperspectral representation, yet its application in cross-domain transfer scenarios remains under-explored. Existing methods, however, still rely on source domain annotations and are susceptible to distribution shifts, leading to degraded generalization performance in the target domain. To address this, this paper proposes a self-supervised cross-domain transfer framework that learns transferable spectral-spatial joint representations without source labels and achieves efficient adaptation under few samples in the target domain. During the self-supervised pre-training phase, a Spatial-Spectral Transformer (S²Former) module is designed. It adopts a dual-branch spatial-spectral transformer and introduces a bidirectional cross-attention mechanism to achieve spectral-spatial collaborative modeling: the spatial branch enhances structural awareness through random masking, while the spectral branch captures fine-grained differences. Both branches mutually guide each other to improve semantic consistency. We further propose a Frequency Domain Constraint (FDC) to maintain frequency-domain consistency through real Fast Fourier Transform (rFFT) and high-frequency magnitude loss, thereby enhancing the model's capability to discern fine details and boundaries. During the fine-tuning phase, we introduce a Diffusion-Aligned Fine-tuning (DAFT) distillation mechanism. This aligns semantic evolution trajectories through a teacher-student structure, enabling robust transfer learning under low-label conditions. Experimental results demonstrate stable classification performance and strong cross-domain adaptability across four hyperspectral datasets, validating the method's effectiveness under resource-constrained conditions.
\end{abstract}


\end{frontmatter}


\section{Introduction}\label{Intro}

Characterized by reflectance acquisition in numerous consecutive narrow spectral bands \citep{Kumar2024HSISurvey}, hyperspectral imagery (HSI) provides dense spectral-spatial information that has been extensively leveraged in precision agriculture, mineral exploration, environmental monitoring, and diverse remote sensing domains \citep{wang2023advances,SONG2025103285,AHMAD2025130428}. 

In recent, deep learning has driven profound progress in hyperspectral image classification (HSIC) \citep{chen2014deep,PAOLETTI2019279}, with convolutional neural networks (CNNs) \citep{makantasis2015deep}, 3D-CNNs \citep{Zhang2020Tree3DCNN}, and Transformer architectures \citep{FANG2025104308} exhibiting strong feature representational power under sufficient annotations. For instance, SSTN \citep{zhong2022spectral} boosts performance through self-attention and multi-scale integration in supervised settings \citep{shi2022explainable}. Nevertheless, these supervised approaches are strongly reliant on massive labeled samples\citep{polewski2016combining,wang2022new}. The prohibitive annotation costs and low efficiency consequently restrict their practical deployment in data-scarce environment \citep{zhao2022superpixel}.

With rising interest in reducing annotation reliance, self-supervised learning (SSL) has gained prominence \citep{chen2023discriminative,khemiri2025deepcontrastive}, with two primary paradigms emerging: masked modeling and contrastive learning \citep{liang2025supercot}. The former includes approaches such as SS-MAE \citep{lin2023ssmae} , which randomly masks spatial and spectral regions to force contextual representation learning; $\text{S}^2\text{HM}^2$ \citep{tu2024s2hm2}, incorporating hierarchical masking combined with spectral angular error; and SSLSM \citep{liu2023self}, simulating spectral correlations through band prediction. The latter encompasses S3L \citep{guo2024s3l}, integrating multi-view consistent Transformers; self-supervised low-rank priors \citep{10943654}, compressing redundancy through spectral low-rank priors; $\text{S}^4\text{L-FSC}$ \citep{chen2025spectral} and ES2FL \citep{liu2022es2fl}, employing class-aware contrastive learning and multiple augmentation strategies respectively to enhance discriminative capability and intra-class consistency. These methods have achieved remarkable progress in homogeneous domain scenarios\citep{Wambugu2021HSICReview}.

However, significant distribution discrepancies (domain shifts) arise between different hyperspectral data sources due to varying imaging conditions, sensor specifications, and geographical contexts, severely compromising model generalization in new domains\citep{wang2025test,huang2020deep}. Traditional domain adaptation (DA) methods typically achieve unsupervised transfer via sample reweighting, feature alignment, or adversarial training, as exemplified by MMD \citep{long2015learning}, and DANN \citep{ganin2016domain}. 
These adaptation paradigms are gaining traction in HSIC for handling distribution shifts across multi-source scenaries \citep{Zhou_2023_CVPR}. For instance, MLUDA achieves joint alignment across pixel, feature, and logical levels, enhancing transfer performance through guided filtering and cross-attention \citep{Wu2025MultiLevelDA}; S4DL \citep{feng2025s4dl} introduces spectral-spatial decoupling and reversible structures to model inter-domain shifts. Open-Set DA algorithm addresses unknown class challenges by incorporating weighted generative adversarial networks with dynamic thresholding strategies \citep{Zhang2025OpenSetDA}; DCFSL \citep{li2022deep} minimizes feature space discrepancies between source and target domains to enhance transferability; FDFSL \citep{qin2024cross} improves robustness in cross-domain few-shot learning through feature alignment strategies; HyMuT \citep{liu2024hybrid} employs multimodal fusion combining spectral and spatial features to enhance model adaptability in complex cross-domain scenarios. Although these approaches advance transfer performance, they still suffer from insufficient semantic modeling and limited transferability under extremely low-label conditions\citep{2017MugNet}.

Emerging solutions hybridize reconstruction mechanisms, contrastive learning, and distillation strategies for hyperspectral cross-domain transfer\citep{chen2023discriminative}. DEMAE \citep{10639453} unifies masked modeling and diffusion processes, demonstrating strong discriminative capability and generalization performance under few-shot conditions through joint pre-training of reconstruction and diffusion. MSDDA \citep{10620320} introduces mask-assisted reconstruction and semantic consistency distillation, improving feature alignment stability and discriminative power; CDnet \citep{lee2022self} constructs positive--negative sample pairs via cross-image contrastive learning, achieving preliminary self-supervised alignment in multi-source scenarios, effectively mitigating the label scarcity problem in target domain. Additionally, CTF \citep{xi2024ctf} integrates semi-supervised and few-shot learning to enhance model performance under scarce label conditions. Despite these advances, feature alignment robustness and representation transferability persist as bottlenecks in complex cross-domain environments \citep{li2023supervised,thota2021contrastive}.

Therefore, this paper proposes a self-supervised cross-domain transfer framework for HSIC that learns transferable spectral-spatial joint representations independent of source annotations, addressing key limitations of label dependency and generalization gaps in cross-domain HSIC \citep{10191249,zhang2023generic}. It integrates self-supervised spectral–spatial synergy with diffusion-based semantic alignment for effective few-shot adaptation. Experimental evaluations confirm notable improvements in boundary sensitivity, semantic consistency, and cross-domain generalization.

The principal innovations of this paper are stated as follows:

\begin{enumerate}[label=(\arabic*), itemindent=1em]
    \setlist[enumerate, 1]{itemindent=1.7em}  
    \item A self-supervised cross-domain classification framework is proposed, which combines masked modeling with frequency-domain awareness during pre-training, allowing it to learn transferable spectral-spatial representations for lightweight adaptation with minimal target samples.
 
    \item A Spatial-Spectral Transformer (S$^2$Former) module is designed, which integrates spatial masking and spectral guidance in a dual-branch Transformer architecture, using bidirectional cross-attention for collaborative spectral--spatial modeling.

    \item A Frequency Domain Constraint (FDC) is introduced, which leverages real Fast Fourier Transform (rFFT) and frequency band masking to construct a high-frequency mask loss, which enhances the model's capacity to capture spectral details and boosts its cross-domain discriminability.

    \item A Diffusion-Aligned Fine-tuning (DAFT) distillation mechanism is developed to construct a Diffusion trajectory aggregation loss, guiding the student model to learn the teacher's semantic evolution path in the target domain and mitigate semantic drift.

\end{enumerate}

The subsequent sections are arranged in the following order. Section~\ref{method} details the proposed framework. Section~\ref{experiments} evaluates the method on several public datasets. Section~\ref{conclusion} provides conclusions and future directions.

\section{Methodology}\label{method}

\begin{figure}[H]
    \centering
    \includegraphics[width=\textwidth]{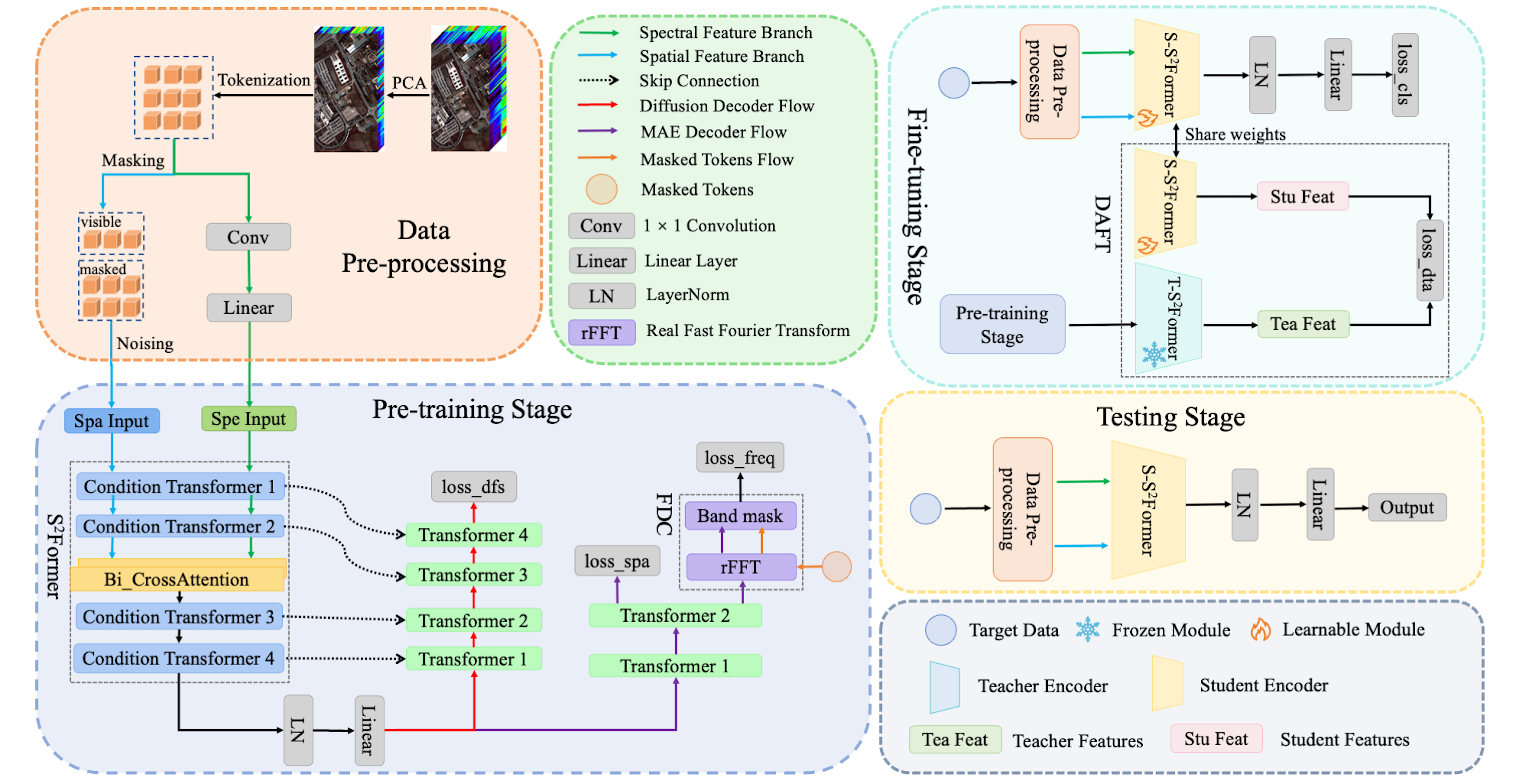}
    \caption{Overview of the proposed self-supervised cross-domain framework, encompassing four stages: data pre-processing, self-supervised pre-training, few-shot cross-domain fine-tuning, and testing.}
    \label{fig:figure1}
\end{figure}

As illustrated in Figure \ref{fig:figure1}, our framework comprises four coordinated stages of spectral-spatial processing. Its core includes three specialized modules: S²Former for joint spatial-spectral dual-stream modeling, FDC for spectral enhancement, and DAFT for cross-domain adaptation.

\begin{figure}[H]
    \centering
    \includegraphics[width=0.4\textwidth]{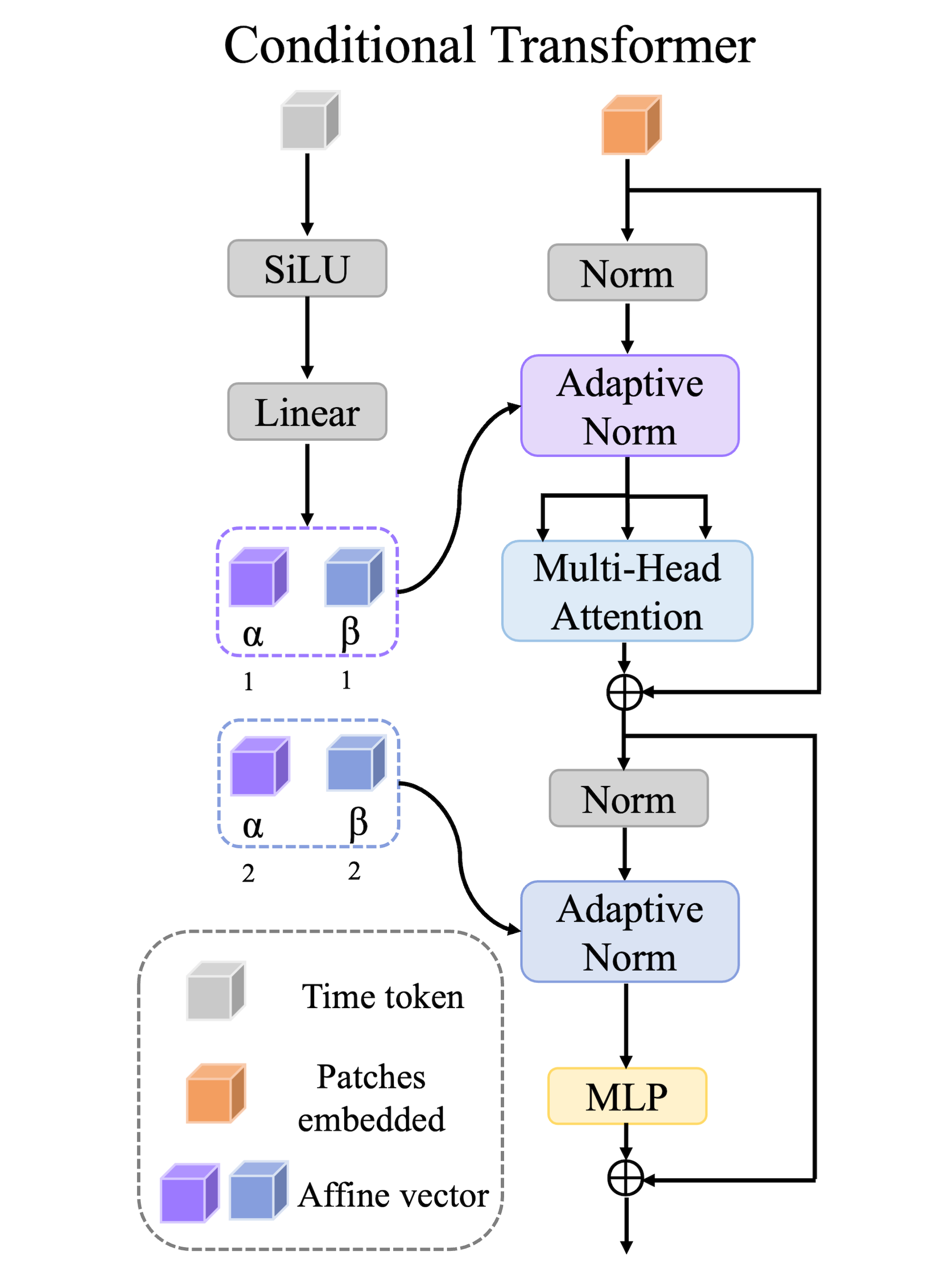}
    \caption{Condition transformer model \citep{10639453}}
    \label{fig:figure2}
\end{figure}

Building upon DEMAE's conditional transformer and noise processing paradigm, our framework enhances temporal token denoising through a learnable normalization layer (Adaptive Norm) with radial parametric generation ($\alpha$/$\beta$), as shown in Figure \ref{fig:figure2}. By integrating Markov-based noise injection with an inter-layer skip connected transformer decoder, the model achieves robust noise immunity and feature recovery.

\subsection{S²Former (Spatial-Spectral Transformer)}

S²Former employs a four-layer conditional transformer architecture. The first two layers use parallel spatial-spectral dual-branch to capture local spatial context and spectral distribution characteristics, respectively, while the intermediate layer applies bidirectional cross-attention mechanism for deep cross-modal complementarity.

\subsubsection{Input modeling and encoding path}

As depicted in Figure \ref{fig:figure1}, the pre-processing stage applies principal component analysis (PCA) to produce \( \mathbf{X} \in \mathbb{R}^{H \times W \times C} \), followed by segmentation into \( p \times p \) patches that are rearranged into a token sequence  $\mathbf{S}$ for mini-batch training:

\begin{equation}
\mathbf{S} = \left[\mathbf{s}_1, \mathbf{s}_2, \dots, \mathbf{s}_N\right]^\top \in \mathbb{R}^{B \times N \times C}, \quad N = \frac{H \cdot W}{p^2},
\end{equation}

\noindent where \( H \), \( W \) and \( C \) denote the height, width, and spectral dimension of \(\mathbf{X}\), respectively. \( B \) denotes the batch size. \( N \) represents the number of tokens, with each token \( \mathbf{s}_n (n=1,\cdots,N) \) corresponding to a \( p \times p \) image patch. 

\begin{figure}[H]
    \centering
    \includegraphics[width=0.7\textwidth]{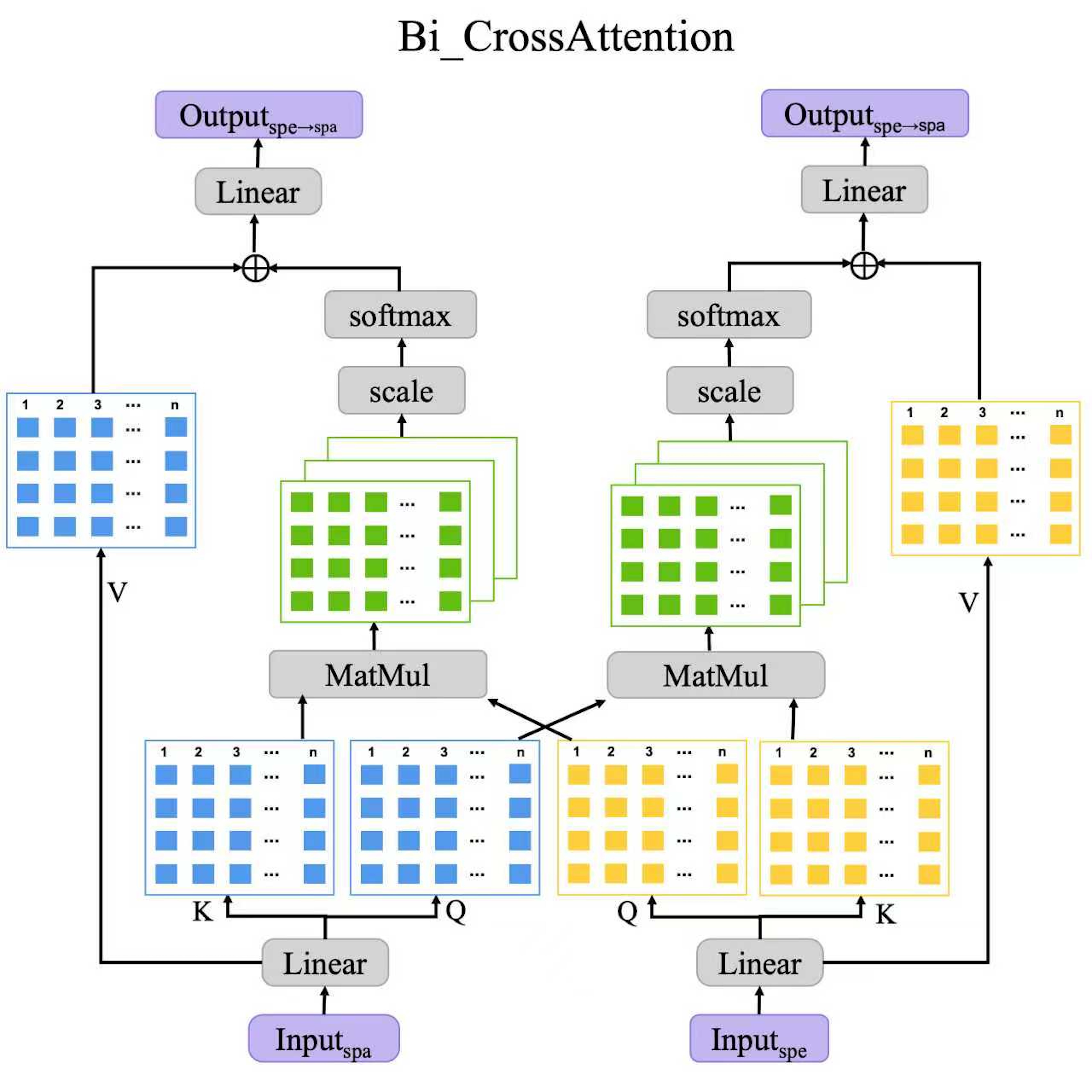}
    \caption{Bidirectional Cross-Attention Module}
    \label{fig:figure3}
\end{figure}

Next, two decoupled encoding branches are constructed:

\vspace{3pt}

\textbf{1) Spatial branch}

The spatial branch randomly masks each token at a preset ratio, dividing them into \( X_{\text{spa}}^{\text{vis}} \) and \( X_{\text{spa}}^{\text{mask}} \), then adds the diffusion time encoding \( T \in \mathbb{R}^{1 \times D} \) to \( X_{\text{spa}}^{\text{vis}} \).

\begin{equation}
\mathbf{Z}_{\text{spa}}^{(0)} = \mathbf{X}_{\text{spa}}^{\text{vis}} \cdot \mathbf{W}_{\text{spa}} + \mathbf{T}, \quad 
\mathbf{W}_{\text{spa}} \in \mathbb{R}^{C \times D},
\end{equation}

\noindent where \( \mathbf{Z}_{\text{spa}}^{(0)} \in \mathbb{R}^{B \times N_{\text{vis}} \times D} \) is the spatial feature sequence, with $N_{\text{vis}}$ and \( D \) representing the visible token count and feature dimension, and \( \mathbf{W}_{\text{spa}} \) the spatial embedding matrix. 

The sequence \(\mathbf{Z}_{\text{spa}}^{(0)} \) undergoes hierarchical feature transformation through two conditional transformer encoders. 
\begin{equation}
\mathbf{Z}_{\text{spa}}^{(l)} = \mathrm{CTransformer}\!\big(\mathbf{Z}_{\text{spa}}^{(l-1)} \big)  \quad (l = 1, 2),
\end{equation}

\noindent where \( \text{CTransformer}(\cdot) \) denotes the conditional transformer, \( \mathbf{Z}_{\text{spa}}^{(l)} \) the current layer output, and \( \mathbf{Z}_{\text{spa}}^{(l-1)} \) the previous layer input.

\vspace{3pt}
\textbf{2) Spectral branch}

The spectral branch applies 1D convolution along the spectral dimension to process mini-batched tokens \( \mathbf{S} \).

\begin{equation}
\hat{\mathbf{X}}_{\text{spec}} = \text{Conv1D}(\mathbf{S}; \mathbf{W}_{\text{conv}}),
\end{equation}

\noindent where $W_{\text{conv}} \in \mathbb{R}^{K \times C}$ is the convolution kernel with size $K$. 

Subsequent linear projection maps each token's spectral features to the target feature space.

\begin{equation}
\mathbf{X}_{\text{spec}} = \text{Linear}(\hat{\mathbf{X}}_{\text{spec}}; \mathbf{W}_{\text{spec}}),
\end{equation}
\noindent where $\mathbf{W}_{\text{spec}} \in \mathbb{R}^{C \times D}$ is the projection matrix.

Diffusion time encoding $\mathbf{T} \in \mathbb{R}^{1 \times D}$ is then added into the spectral features, yielding the final spectral feature sequence $\mathbf{Z}_{\text{spec}}^{(0)} \in \mathbb{R}^{B \times N \times D}$:
\begin{equation}
\mathbf{Z}_{\text{spec}}^{(0)} = \mathbf{X}_{\text{spec}} + \mathbf{T}.
\end{equation}

Spectral features pass through two conditional transformer layers, evolving from \( \mathbf{Z}_{\text{spec}}^{(0)} \) into \(\mathbf{Z}_{\text{spec}}^{(2)} \).

\subsubsection{Bidirectional cross-attention}

Our encoder utilizes bidirectional cross-attention (Figure \ref{fig:figure3}) for dual interaction and cross-modal fusion.

\vspace{3pt}
\textbf{1) Spectral-to-spatial}

First, spectral tokens $\mathbf{Z}_{\text{spec}}^{(2)} \in \mathbb{R}^{B \times N \times D}$ are averaged by:
\begin{equation}
\bar{\mathbf{Z}}_{\text{spec}} = \frac{1}{N} \sum_{i=1}^{N} \mathbf{Z}_{\text{spec}}^{{(2)}{(i)}}. 
\end{equation}

Proceed to form the cross-attention inputs:
\begin{equation}
\mathbf{Q} = \mathbf{Z}_{\text{spa}}^{(2)} \cdot \mathbf{W}_Q, \quad 
\mathbf{K} = \bar{\mathbf{Z}}_{\text{spec}} \cdot \mathbf{W}_K, \quad 
\mathbf{V} = \bar{\mathbf{Z}}_{\text{spec}} \cdot \mathbf{W}_V,
\end{equation}

\noindent where $\mathbf{W}_Q, \mathbf{W}_K, \mathbf{W}_V \in \mathbb{R}^{D \times d}$ are projection matrices, and $d$ is the attention head dimension.

The frequency-guided spatial attention is defined as:
\begin{equation}
\text{Attn}_{\text{spa}} = \text{Softmax}\left( \frac{\mathbf{Q} \mathbf{K}^\top}{\sqrt{d}}. \right) \mathbf{V}.
\end{equation}

Then, the fusion residual is expressed as:
\begin{equation}
\hat{\mathbf{Z}}_{\text{spa}} = \mathbf{Z}_{\text{spa}}^{(2)} + \text{Attn}_{\text{spa}}.
\end{equation}

This spectral-guided mechanism directs the spatial pathway to focus on spectral-sensitive regions, achieving complementary feature enhancement.

\vspace{3pt}
\textbf{2) Spatial-to-spectral}

Similarly, mean pooling aggregates the spatial representation:

\begin{equation}
\bar{\mathbf{Z}}_{\text{spa}} = \frac{1}{N_{\text{vis}}} \sum_{i=1}^{N_{\text{vis}}} \mathbf{Z}_{\text{spa}}^{{(2)}{(i)}}
\end{equation}

Construct spectral attention inputs:
\begin{equation}
\mathbf{Q} = \mathbf{Z}_{\text{spec}}^{(2)} \cdot \mathbf{W}_Q', \quad 
\mathbf{K} = \bar{\mathbf{Z}}_{\text{spa}} \cdot \mathbf{W}_K', \quad 
\mathbf{V} = \bar{\mathbf{Z}}_{\text{spa}} \cdot \mathbf{W}_V',
\end{equation}

\noindent where $\mathbf{W}_Q', \mathbf{W}_K', \mathbf{W}_V' \in \mathbb{R}^{D \times d}$ are the attention mapping matrices.

Thus, yielding the spatial-guided spectral attention output:

\begin{equation}
\text{Attn}_{\text{spec}} = \text{Softmax}\left( \frac{\mathbf{Q} \mathbf{K}^\top}{\sqrt{d}} \right) \mathbf{V}.
\end{equation}

The fusion residual in this direction is represented as:
\begin{equation}
\hat{\mathbf{Z}}_{\text{spec}} = \mathbf{Z}_{\text{spec}}^{(2)} + \text{Attn}_{\text{spec}}.
\end{equation}

The cross-attention refined spatial features \(\hat{\mathbf{Z}}_{\text{spa}}\) and spectral features \(\hat{\mathbf{Z}}_{\text{spec}}\) are concatenated and linearly transformation into a unified representation:

\begin{equation}
\hat{\mathbf{Z}}_{\text{fuse}} = \mathrm{Linear}\!\left( \mathrm{concat}\!\left[\hat{\mathbf{Z}}_{\text{spa}},\, \hat{\mathbf{Z}}_{\text{spec}} \right] \right).
\end{equation}

Subsequently, \(\hat{\mathbf{Z}}_{\text{fuse}}\) undergoes feature refinement via two conditional transformer layers, followed by LayerNorm and linear layer, to generate enhanced features \(\mathbf{Z}_{\text{fuse}}\) for subsequent tasks.

\subsection{Frequency-Domain Constraint (FDC)}\label{FDC}

To improve the model's ability to capture fine-grained spectral features in hyperspectral images, this paper introduces a Frequency-Domain Constraint (FDC) during the self-supervised pre-training phase,as shown in Figure \ref{fig:figure1}. This module employs explicit spectral-domain supervision to guide the model toward focusing on high-frequency components at masked locations, thereby enhancing its representational capacity and cross-domain generalization performance.

The reconstruction decoder initializes the sequence by placing \(\mathbf{Z}_{\text{fuse}}\) at visible positions and learnable tokens $\mathbf{M}_{\text{mask}}^{\text{init}}$ at masked regions. After adding shared positional embedding $\mathbf{P}$ to all tokens to form $\mathbf{X}_{\text{dec}}^{(0)}$, a two-layer transformer decoder processes it to generate the final output $\mathbf{X}_{\text{dec}}$, where the reconstructed features $\hat{\mathbf{X}}_{\text{spa}}^{\text{mask}}$ are extracted from masked positions.

To evaluate spectral restoration in the frequency domain, we apply the rFFT to both reconstructed \( \hat{\mathbf{X}}_{\text{spa}}^{\text{mask}} \) and ground-truth \( \mathbf{X}_{\text{spa}}^{\text{mask}} \):

\begin{equation}
\tilde{\mathbf{X}}_{\text{freq},n} = \mathcal{F}_r(\mathbf{X}_{\text{spa},n}^{\text{mask}}), \quad \hat{\tilde{\mathbf{X}}}_{\text{freq},n} = \mathcal{F}_r(\hat{\mathbf{X}}_{\text{spa},n}^{\text{mask}}), \quad \forall n \in \mathcal{M},
\end{equation}

\noindent where \( \mathcal{M} \in \{1, \dots, M\} \) is the masked token set, with \( \mathbf{X}_{\text{spa}, n}^{\text{mask}} \) and \( \hat{\mathbf{X}}_{\text{spa}, n}^{\text{mask}} \) being the ground-truth and  predicted value at the \( n \)-th masked position. The rFFT operation $\mathcal{F}_r(\cdot)$ along the channel dimension produces a complex spectrum of dimension $C' = \lfloor C/2 \rfloor + 1$.

To emphasize high-frequency components, we employ a frequency band mask $\mathbf{m} \in \{0,1\}^{C'}$ defined as:

\begin{equation}
\mathbf{m}_i =
\begin{cases}
1, & \text{if } i > \tau C' \\
0, & \text{otherwise}
\end{cases}
\quad \text{with } \tau \in (0, 1),
\end{equation}

\noindent where $\tau$ controls the high-frequency band's start ($\tau = 0.3$ ), $i \in \{0, 1, \dots, C'-1\}$ denotes the frequency channel index. $\mathbf{m}_i = 1$ selects high-frequency components for loss calculation, while $\mathbf{m}_i = 0$ excludes low-frequency channels.

\subsection{Diffusion-Aligned Fine-tuning (DAFT)}

This paper proposes Diffusion-Aligned Fine-tuning (DAFT), using a frozen source-domain teacher to guide a target-domain student through a tripartite joint loss (classification, timestep alignment, and distillation consistency) for stable few-shot transfer (Figure \ref{fig:figure1}).

An input sample \( \mathbf{X} \) is partitioned into tokens \( S \), then perturbed with diffusion noise at a random timestep \( t \in \mathcal{K}, \mathcal{K} = \{1, 2, \ldots, T_S\}  \), where \( T_S \) represents the total diffusion steps.

\begin{equation}
\mathbf{x}^{(t)} = \sqrt{\bar{\alpha}_t} \cdot \mathbf{x} + \sqrt{1 - \bar{\alpha}_t} \cdot \epsilon, \quad \epsilon \sim N(0, \mathbf{I}),
\end{equation}

\noindent where $\bar{\alpha}_t$ denotes the cumulative retention factor and $\epsilon$ represents standard Gaussian noise, and $\mathbf{x}$ refers to the token-level representation sampled from the token sequence $\mathbf{S}$.

At timestep $t$, the teacher $\mathbf{f}_T$ and student $\mathbf{f}_S$ respectively output features:
\begin{equation}
\mathbf{z}_T^{(t)} = \mathbf{f}_T(\mathbf{x}^{(t)}), \quad \mathbf{z}_S^{(t)} = \mathbf{f}_S(\mathbf{x}^{(t)}),
\end{equation}
 
\noindent where $\mathbf{z}_T^{(t)}$ and $\mathbf{z}_S^{(t)}$ are the class tokens from the teacher and student models at timestep $t$, used for measuring their feature consistency within the diffusion semantic space.

\subsection{Loss functions}

To enhance the model's generalization and cross-domain adaptability, we design multiple loss functions for joint optimization during both pre-training and fine-tuning. 
 
\subsubsection{Pre-training losses}
 
The pre-training phase optimizes the encoder through self-supervision using two primary objective functions.

\vspace{3pt}
1) Signal-guided classification loss

The masked reconstruction branch loss combines spatial and frequency-domain constraint term to guide encoder reconstruction of both spatial structures and spectral features.

The spatial reconstruction term employs a pointwise \( \ell_1 \) loss:
\begin{equation}
\mathcal{L}_{\text{spa}} = \frac{1}{|\mathcal{M}|} \sum_{n \in \mathcal{M}} \left\| \hat{\mathbf{X}}_{\text{spa}, n}^{\text{mask}} - \mathbf{X}_{\text{spa}, n}^{\text{mask}} \right\|_1.
\end{equation}

The frequency-domain reconstruction term is formulated as:
\begin{equation}
\mathcal{L}_{\text{freq}} = \frac{1}{|\mathcal{M}|} \sum_{n \in \mathcal{M}} \left\| \mathbf{m} \odot \left( |\tilde{\mathbf{X}}_{\text{freq},n}| - |\hat{\tilde{\mathbf{X}}}_{\text{freq},n}| \right) \right\|_1,
\end{equation}
\noindent where $|\cdot|$ and $\odot$ denote spectral magnitude and element-wise multiplication, constraining the model to preserve accurate distributions in high-frequency regions and thereby enhancing its sensitivity to fine-grained spectral differences.

The reconstruction loss is given by:
\begin{equation}
\mathcal{L}_{\text{recon}} = \mathcal{L}_{\text{spa}} + \alpha \cdot \mathcal{L}_{\text{freq}},
\end{equation}
\noindent where the hyperparameter $\alpha \in \mathbb{R}^+$ controls the importance of frequency-domain supervision (typically 0.5). 
 
 \vspace{3pt}

2) Diffusion loss (DFS loss)

Depicted in Figure \ref{fig:figure1}, a four-layer diffusion decoder processes \(\mathbf{Z}_{\text{fuse}}\) with encoder skip connections, outputting denoised visible features \(\hat{\mathbf{X}}_{\text{vis}}\) that are evaluated against clean targets  \(\mathbf{X}_{\text{spa}}^{\text{vis}}\) via mean squared error (MSE) loss:
\begin{equation}
\mathcal{L}_{\text{dfs}} =\mathrm{MSE}\!\left(\mathbf{X}_{\text{spa}}^{\text{vis}},\,\hat{\mathbf{X}}_{\text{vis}}\right) =\left\|\hat{\mathbf{X}}_{\text{vis}}-\mathbf{X}_{\text{spa}}^{\text{vis}}\right\|_{2}^{2}.
\end{equation}
 
Overall, the final pre-training loss is:
\begin{equation}
\mathcal{L}_{\text{pretrain}} = \mathcal{L}_{\text{recon}} + \mathcal{L}_{\text{dfs}}.
\end{equation}

The two equally weighted losses balance reconstruction of masked spectral-spatial information with robust modeling of visible region perturbations.

\subsubsection{Fine-tuning losses}

Fine-tuning aims for strong discriminative and source-target consistency in the target domain, using two losses as specified below.
 
\vspace{3pt}
1) Signal-to-Noise Ratio-Enhanced classification loss (SNR-Enhanced loss)
 
To enhance model stability across SNR levels after diffusion perturbations, SNR-Enhanced loss is applied:
\begin{equation}
\mathcal{L}_{\text{cls}} = \mathrm{SNREnhanced}(h(z_S^{(0)}), y, \mathrm{SNR}),
\end{equation}

\noindent where \( h(\mathbf{z}_S^{(0)}) \) represents the student's class token, with \( z_S^{(0)} \) being the class token feature obtained from the diffusion-perturbed input, \( y \) is the ground truth, and \( \mathrm{SNR} \) weights the loss by reducing the contribution of high-SNR (low-noise) samples and increasing that of low-SNR (high-noise) samples.

\vspace{3pt}
2) Diffusion Trajectory Aggregation loss

For cross-timestep semantic alignment, the feature consistency loss at each timestep $t$ as:

\begin{equation}
\mathcal{L}_{cd}^{(t)} = 1 - \cos\left( \mathbf{f}_s^{(t)}, \mathbf{P}\left(\mathbf{f}_t^{(t)}\right) \right),
\end{equation}

\noindent where $\mathbf{f}_s^{(t)} \in \mathbb{R}^{D_s}$ and $\mathbf{f}_t^{(t)} \in \mathbb{R}^{D_t}$ are student and teacher features at $t$, $\mathbf{P}(\cdot)$ is an optional projection for the mismatch of $\mathbf{f}_s^{(t)}$ and $\mathbf{f}_t^{(t)}$, and $\cos(\cdot, \cdot)$ represents cosine similarity.

Consequently, we introduce an aggregated loss to enforce globally consistent alignment of the feature trajectories across the entire time series:

\begin{equation}
\mathcal{L}_{\text{dta}} = \frac{1}{|\mathcal{K}|} \sum_{t \in \mathcal{K}} \left[ 1 - \cos\left( \mathbf{f}_s^{(t)}, \mathbf{P}\left( \mathbf{f}_t^{(t)} \right) \right) \right].
\end{equation}

Overall, the final fine-tuning loss is:
\begin{equation}
    \mathcal{L}_{\text{DAFT}} = \mathcal{L}_{\text{cls}} + \lambda \cdot \mathcal{L}_{\text{dta}}. 
\end{equation}
\noindent where $\lambda$ balances alignment and classification, ensuring the student adapts to the target domain while preserving source domain semantic knowledge. 

\section{Experimental results}\label{experiments}

\subsection{Data description}

For rigorous validation of cross-domain transfer capability in HSIC, we employ four benchmark datasets---Pavia Center (PC), Houston2013 (HU), Pavia University (PU), and Salinas (SA)---with Salinas excluding water vapor absorption bands. The selected datasets demonstrate heterogeneous characteristics across spectral details, spatial resolutions, land cover types, and sensor configurations, ensuring thorough assessment of domain adaptation performance. Key parameters of all datasets (size, band information, and labeled sample counts) are detailed in Table~\ref{tab:dataset}.

\begin{table}[H]
\centering
\caption{Size, band information, and number of labeled samples per class for four datasets.}
\label{tab:dataset}
\renewcommand{\arraystretch}{1}
\setlength{\tabcolsep}{6pt}
\resizebox{\textwidth}{!}{
\begin{tabular}{c|cc|cc|cc|cc}
\hline
\textbf{Datasets} &
\multicolumn{2}{c|}{\textbf{PC}} &
\multicolumn{2}{c|}{\textbf{HU}} &
\multicolumn{2}{c|}{\textbf{PU}} &
\multicolumn{2}{c}{\textbf{SA}} \\
\hline
\textbf{Size and bands} &
\multicolumn{2}{c|}{$1096\times715\times102$} &
\multicolumn{2}{c|}{$512\times217\times204$} &
\multicolumn{2}{c|}{$610\times340\times103$} &
\multicolumn{2}{c}{$349\times1905\times144$} \\
\hline
\textbf{Class} & \textbf{Name} & \textbf{Number} &
\textbf{Name} & \textbf{Number} &
\textbf{Name} & \textbf{Number} &
\textbf{Name} & \textbf{Number} \\
\hline
1 & Water & 824          & Healthy grass        & 1251         & Asphalt             & 6631         & Brocoli green weeds 1 & 2009 \\
2 & Trees & 820          & Stressed grass       & 1254         & Meadows             & 18649        & Brocoli green weeds 2 & 3726 \\
3 & Asphalt & 816        & Synthetic grass      & 697          & Gravel              & 2099         & Fallow                & 1976 \\
4 & Bitumen & 808         & Trees                & 1244         & Trees               & 3064         & Fallow rough plow     & 1394 \\
5 & Bricks & 808         & Soil                 & 1242         & Painted metal sheets & 1345         & Fallow smooth         & 2678 \\
6 & Tiles & 1260          & Water                & 325          & Bare soil           & 5029         & Stubble               & 3959 \\
7 & Shadows & 476        & Residential          & 1268         & Bitumen             & 1330         & Celery                & 3579 \\
8 & Bare soil & 824        & Commercial           & 1244         & Self-Blocking Bricks & 3682         & Grapes untrained      & 11271 \\
9 & Meadows & 820         & Road                 & 1252         & Shadows             & 947          & Soil vineyard develop & 6203 \\
10 &  &             & Highway              & 1227         &                     &              & Corn senesced green weeds & 3278 \\
11 &  &             & Railway              & 1235         &                     &              & Lettuce romaine 4wk   & 1068 \\
12 &  &             & Parking Lot1         & 1233         &                     &              & Lettuce romaine 5wk   & 1927 \\
13 &  &             & Parking Lot2         & 469          &                     &              & Lettuce romaine 6wk   & 916 \\
14 &  &             & Tennis court         & 428          &                     &              & Lettuce romaine 7wk   & 1070 \\
15 &  &             & Running track        & 660          &                     &              & Vineyard untrained    & 7268 \\
16 &  &             &                      &              &                     &              & Vineyard vertical trellis & 1807 \\
\hline
   & Total & 7456 & Total & 15029 & Total & 42776 & Total & 54129 \\
\hline
\end{tabular}
}
\end{table}

Transfer Tasks include the following scenarios:

1) \textbf{PU$\rightarrow$PC:} Transfer from campus scenes (PU) to urban architecture (PC) using homologous ROSIS sensors (1.3m), testing adaptation from regular campus to complex urban scenes.
 
2) \textbf{SA$\rightarrow$HU:} Cross-sensor transfer from agriculture SA (AVIRIS 3.7 m) to urban HU (ITRES CASI-1500 2.5m) with significant domain shifts, verifying robustness across sensor and land cover types.

3) \textbf{HU$\rightarrow$PU:} Reverse transfer from complex urban HU to structured campus PU, evaluating model stabilization under distribution divergence.

4) \textbf{PC$\rightarrow$SA:} Urban-to-agricultural transfer between PC (urban buildings) and SA (agricultural parcels) with distinct feature categories, background structures and spectral dimensions, assessing cross-landscape representation capability.

\subsection{Experimental setup} 


Hyperparameter are set according to dataset-adaptive configurations during both pre-training and fine-tuning phases. Specifically, the PU dataset is trained for 200 epochs in both stages. For the SA dataset, pre-training uses 300 epochs and fine-tuning 250 epochs. The HU dataset uses 300 epochs for pret-raining and 200 for fine-tuning, while the PC dataset adopts 300 and 200 epochs for pre-training and fine-tuning, respectively. Regarding the batch size, all datasets uniformly use a size of 512 during pre-training. In the fine-tuning phase, the batch size is adjusted per dataset: 45 for PU and PC, 80 for SA, and 75 for HU. For all datasets, the initial learning rate during both pre-training and fine-tuning phases is fixed at $1 \times 10^{-3}$, with a masking ratio of 0.75. The noise scale is kept consistent across the two phases, set to $[1 \times 10^{-4}, 2 \times 10^{-2}]$, aligning with the classic settings of the DEMAE method. Additionally, a learning rate decay strategy with a decay factor of 0.99 per epoch is applied to ensure training stability.


The experimental sample configuration and performance evaluation design are implemented as follows: For fine-tuning, we randomly select 5 labeled samples per category for training and use the remainder reserved for testing. Evaluation employs Overall Accuracy (OA), Average Accuracy (AA), and Kappa Coefficient ($\kappa$), with results averaged over 10 independent runs (mean \(pm\) stand deviation). All experiments are executed on an RTX 3080 Ti GPU with PyTorch 1.10.0 and CUDA 11.3.

\subsection{Classification results}

We evaluate cross-domain classification with sparse labeled sample on the aforementioned four transfer tasks. Compared methods include: fully supervised SSTN\citep{zhong2022spectral}; few-shot semi-supervised CTF\citep{xi2024ctf}; self-supervised DEMAE\citep{10639453}; and three cross-domain methods DCFSL\citep{li2022deep}, FDFSL\citep{qin2024cross}, HyMuT\citep{liu2024hybrid}.

\begin{figure}[htbp]
    
    \newcommand{\colw}{0.17\textwidth}
    \newcommand{\colsep}{-0.005\textwidth}
    \raggedright
    \hspace*{-2.6em}
    \begin{minipage}{1.3\textwidth}

        \begin{minipage}[t]{\colw}\centering
            \includegraphics[width=\linewidth]{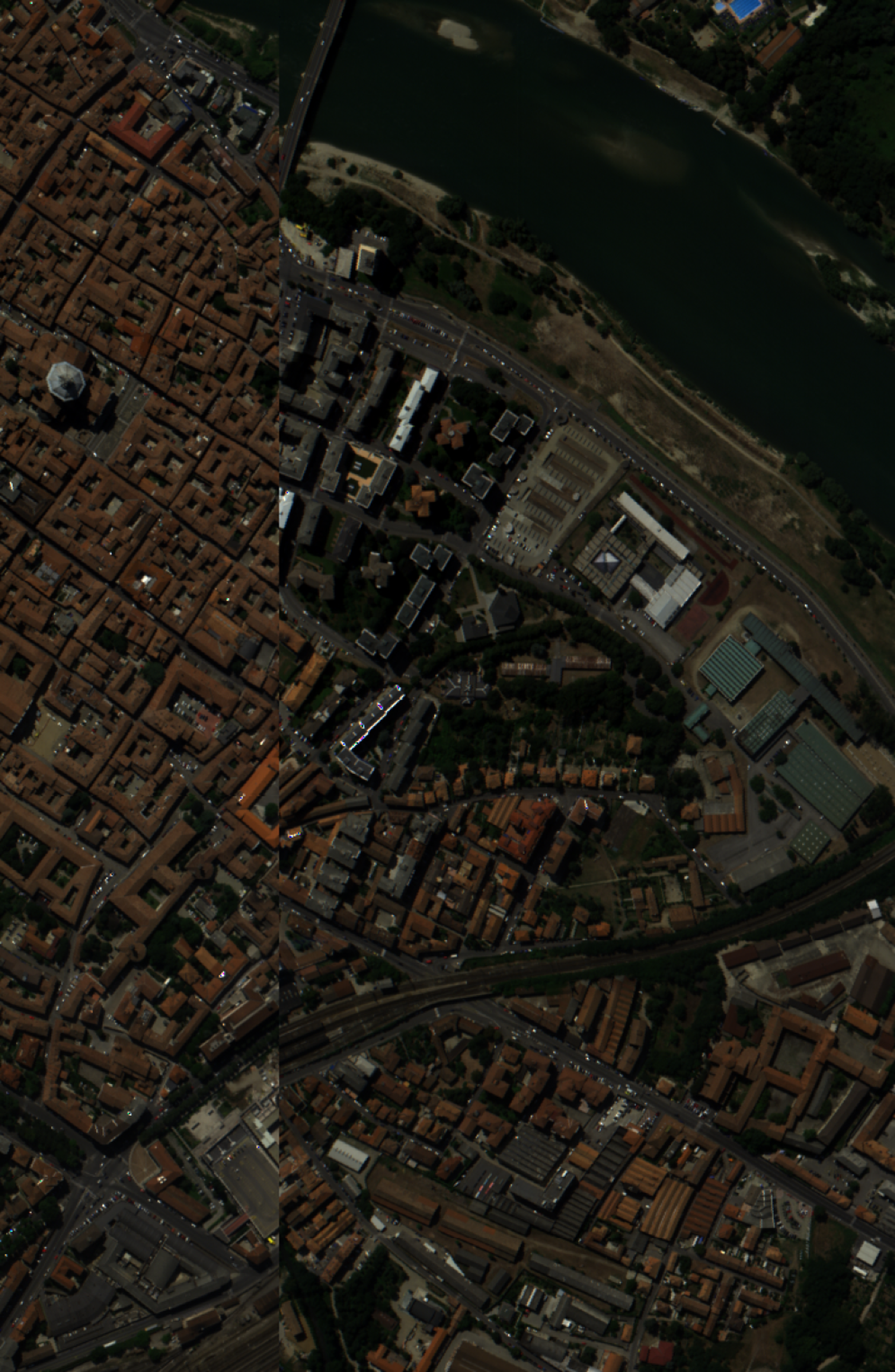}\\[0.2em]
            \tiny(a)
        \end{minipage}\hspace{\colsep}
        \begin{minipage}[t]{\colw}\centering
            \includegraphics[width=\linewidth]{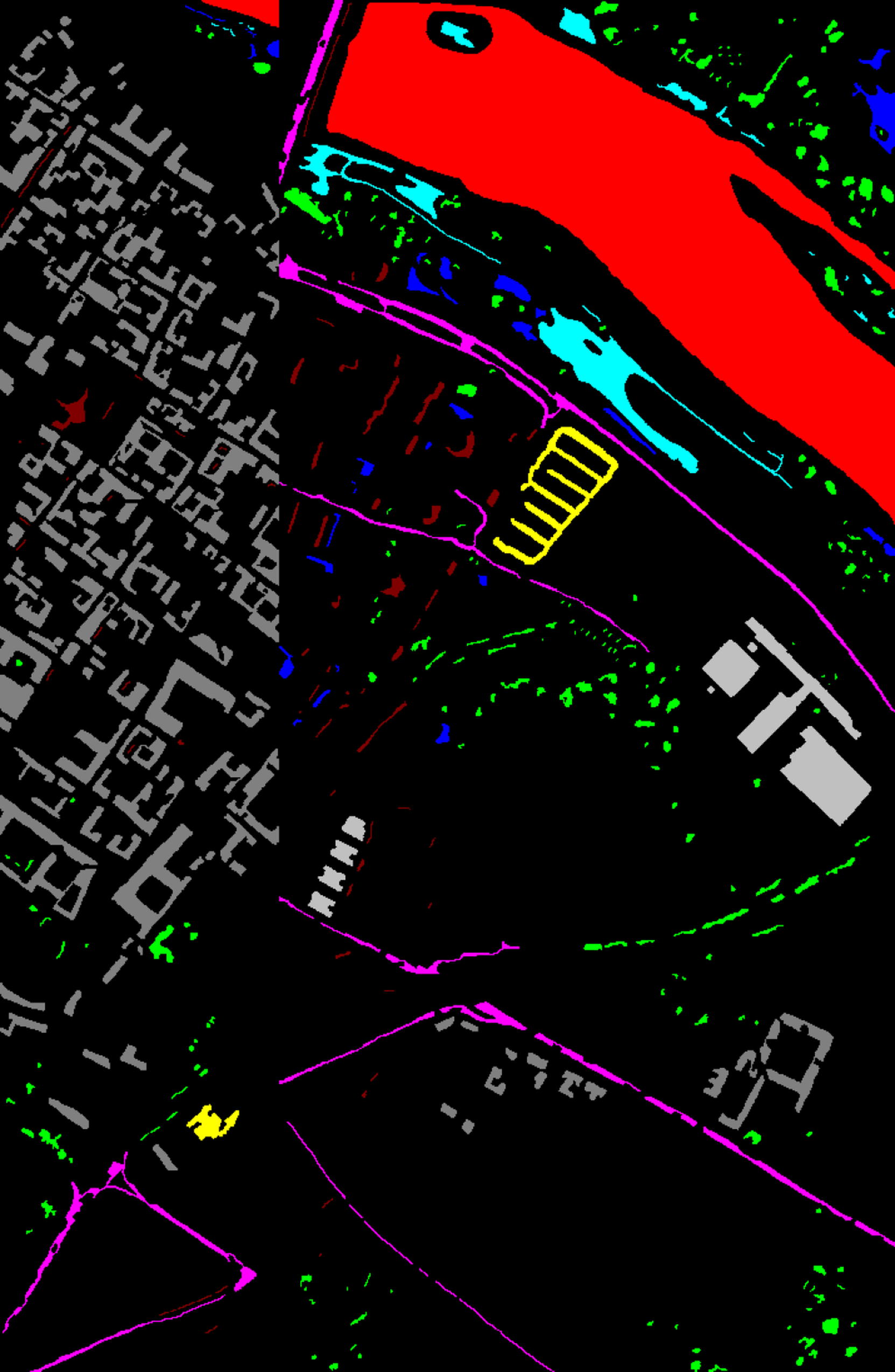}\\[0.2em]
            \tiny(b)
        \end{minipage}\hspace{\colsep}
        \begin{minipage}[t]{\colw}\centering
            \includegraphics[width=\linewidth]{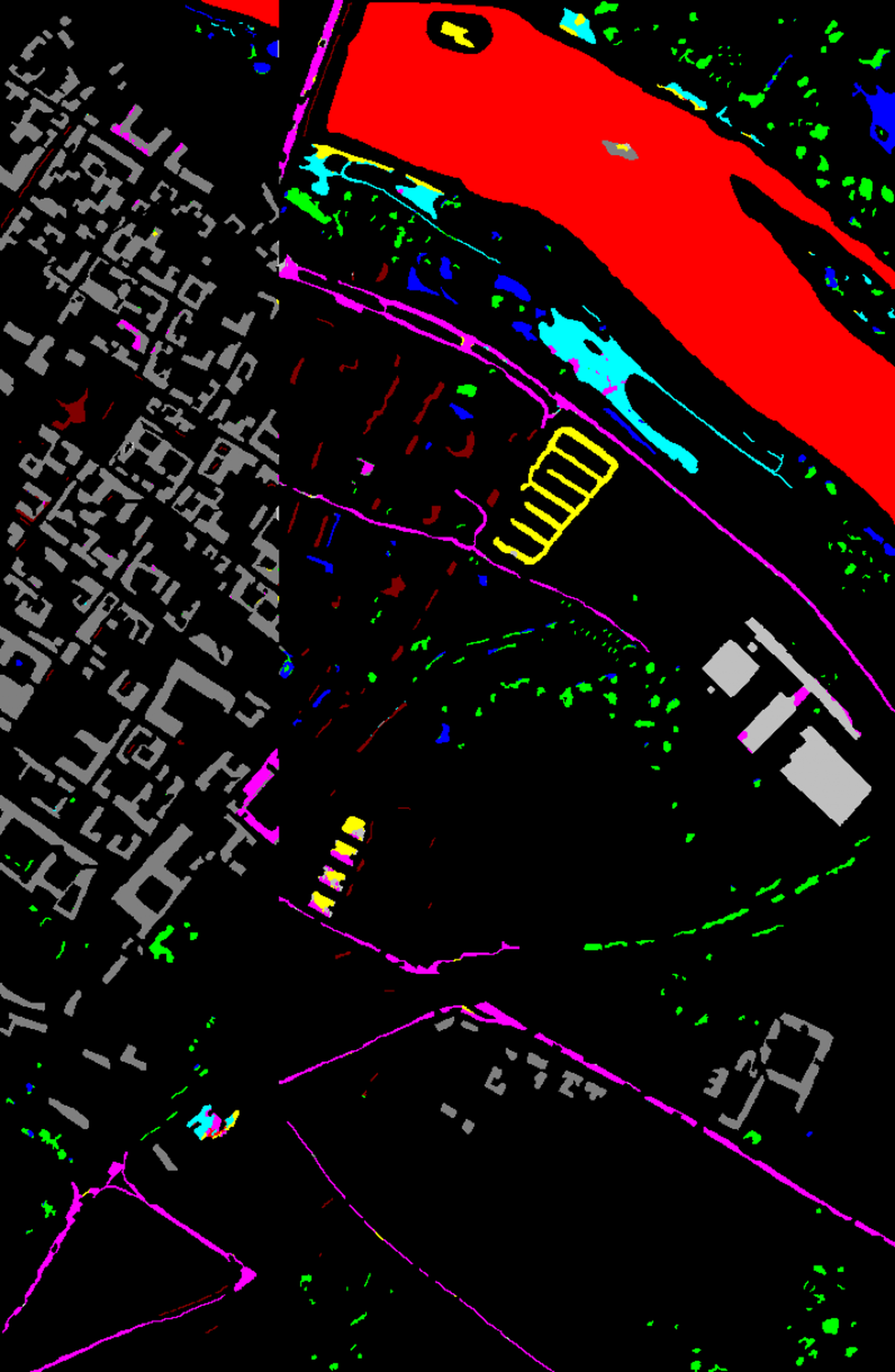}\\[0.2em]
            \tiny(c)
        \end{minipage}\hspace{\colsep}
        \begin{minipage}[t]{\colw}\centering
            \includegraphics[width=\linewidth]{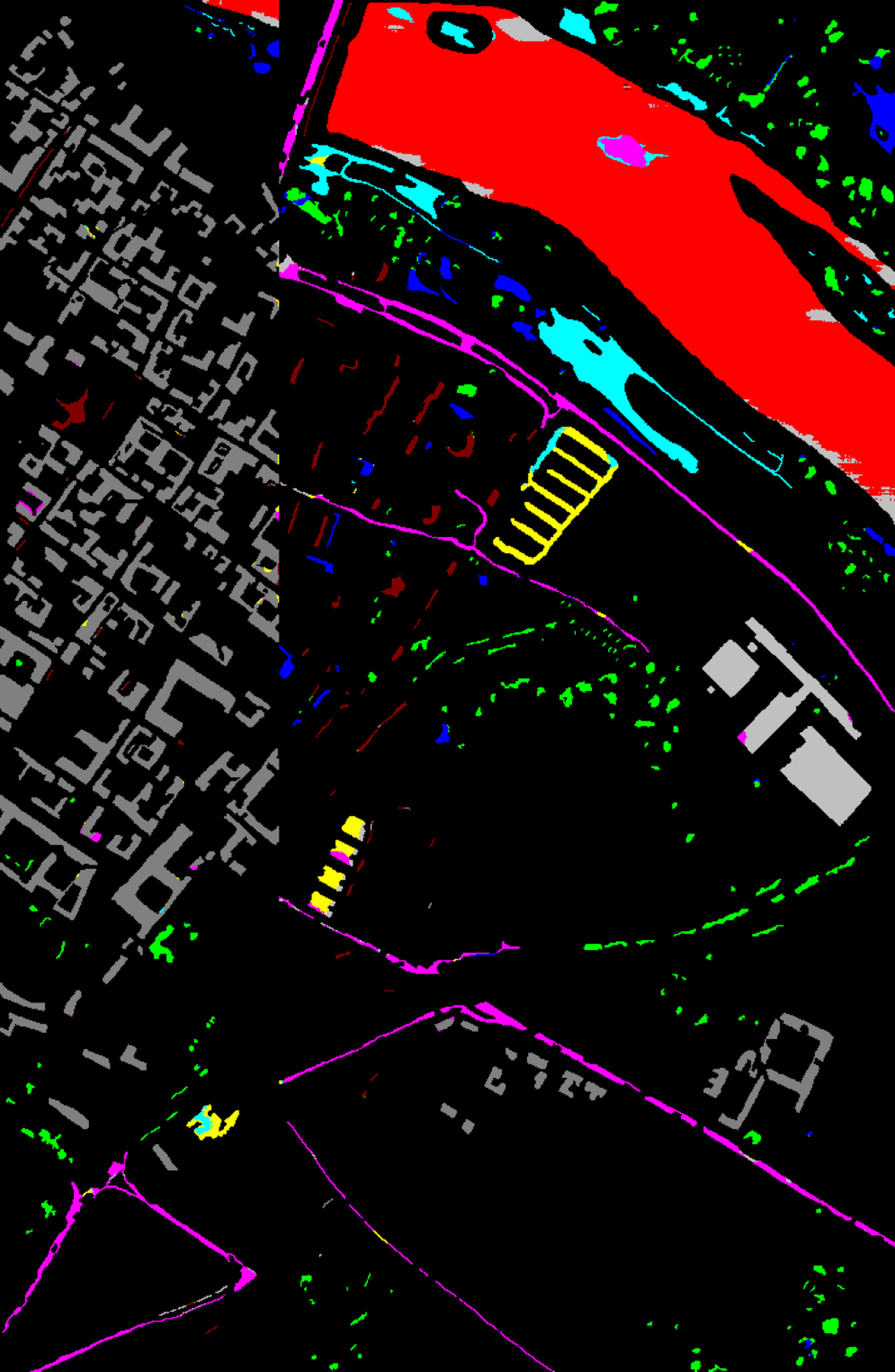}\\[0.2em]
            \tiny(d)
        \end{minipage}\hspace{\colsep}
        \begin{minipage}[t]{\colw}\centering
            \includegraphics[width=\linewidth]{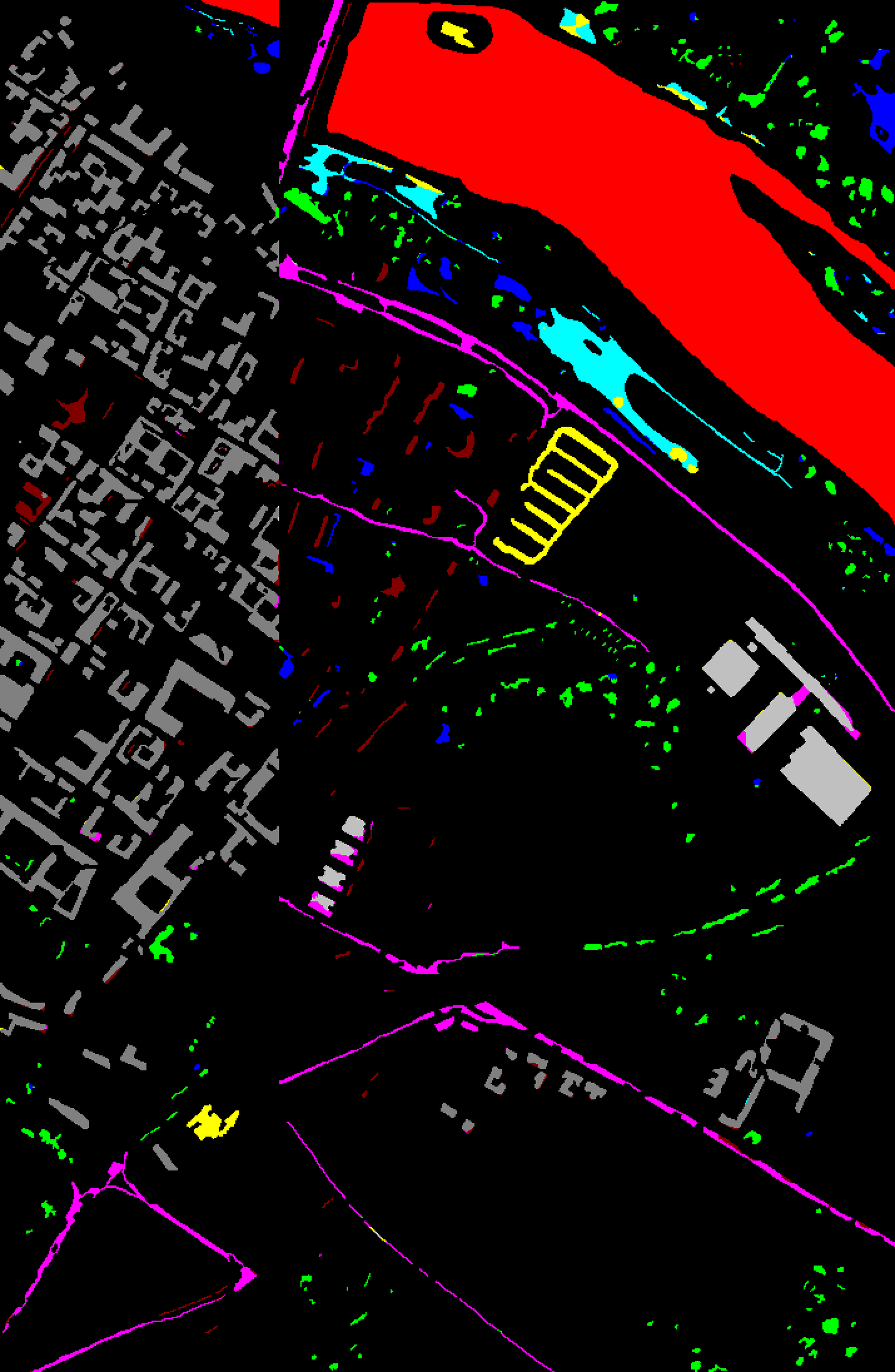}\\[0.2em]
            \tiny(e)
        \end{minipage}
    
        \vspace{0.8em}
    
        \begin{minipage}[t]{\colw}\centering
            \includegraphics[width=\linewidth]{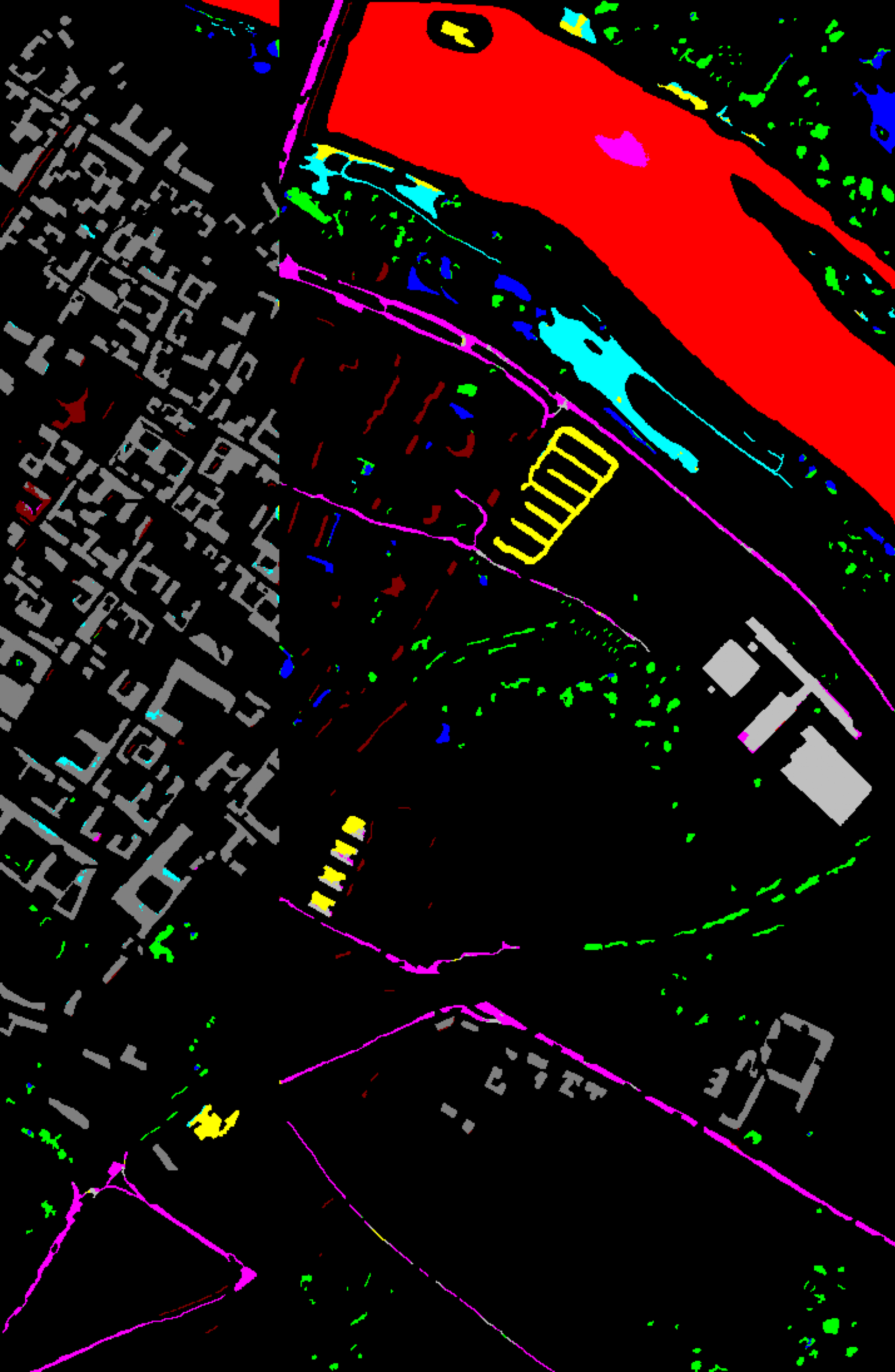}\\[0.2em]
            \tiny(f)
        \end{minipage}\hspace{\colsep}
        \begin{minipage}[t]{\colw}\centering
            \includegraphics[width=\linewidth]{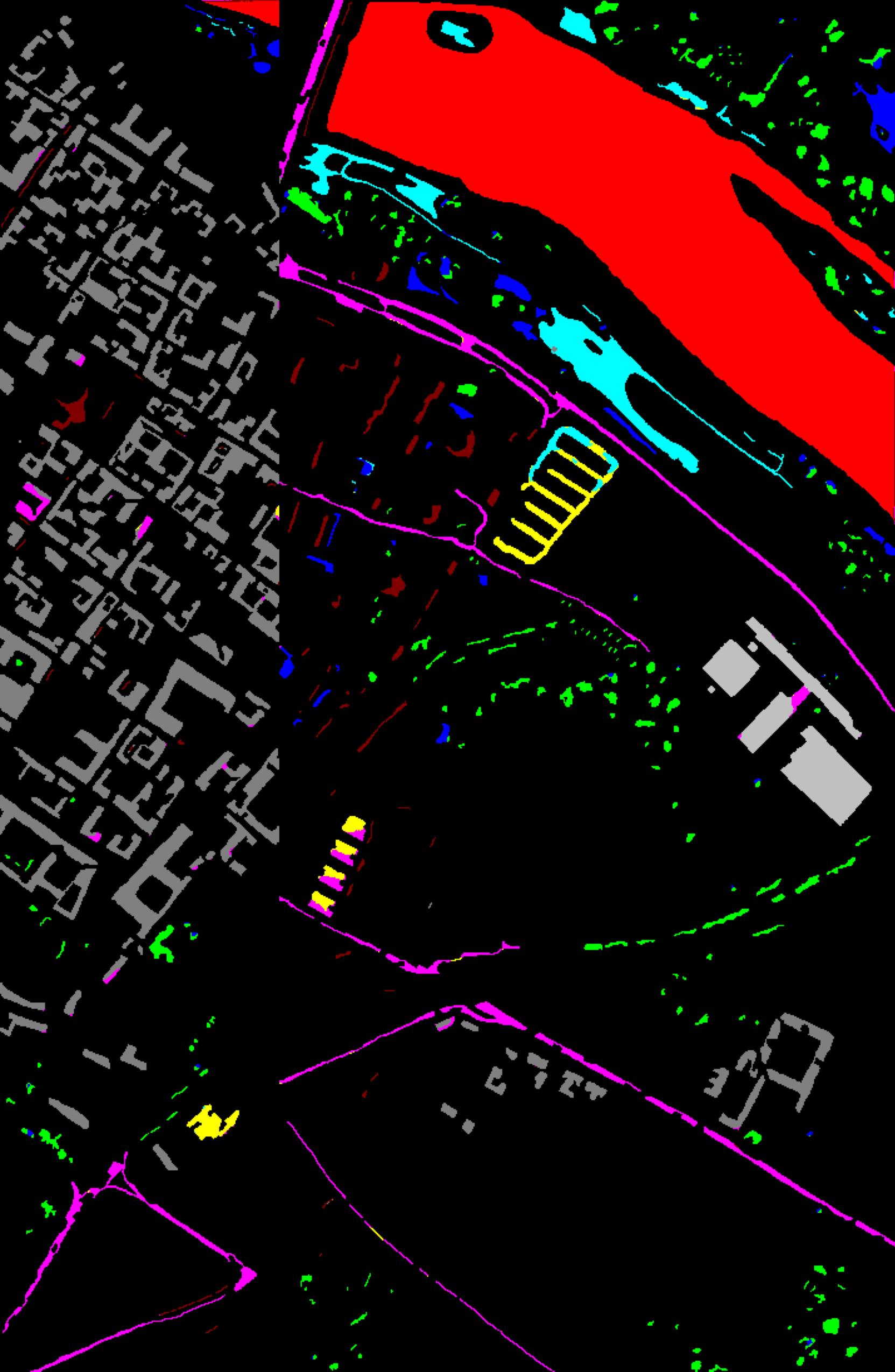}\\[0.2em]
            \tiny(g)
        \end{minipage}\hspace{\colsep}
        \begin{minipage}[t]{\colw}\centering
            \includegraphics[width=\linewidth]{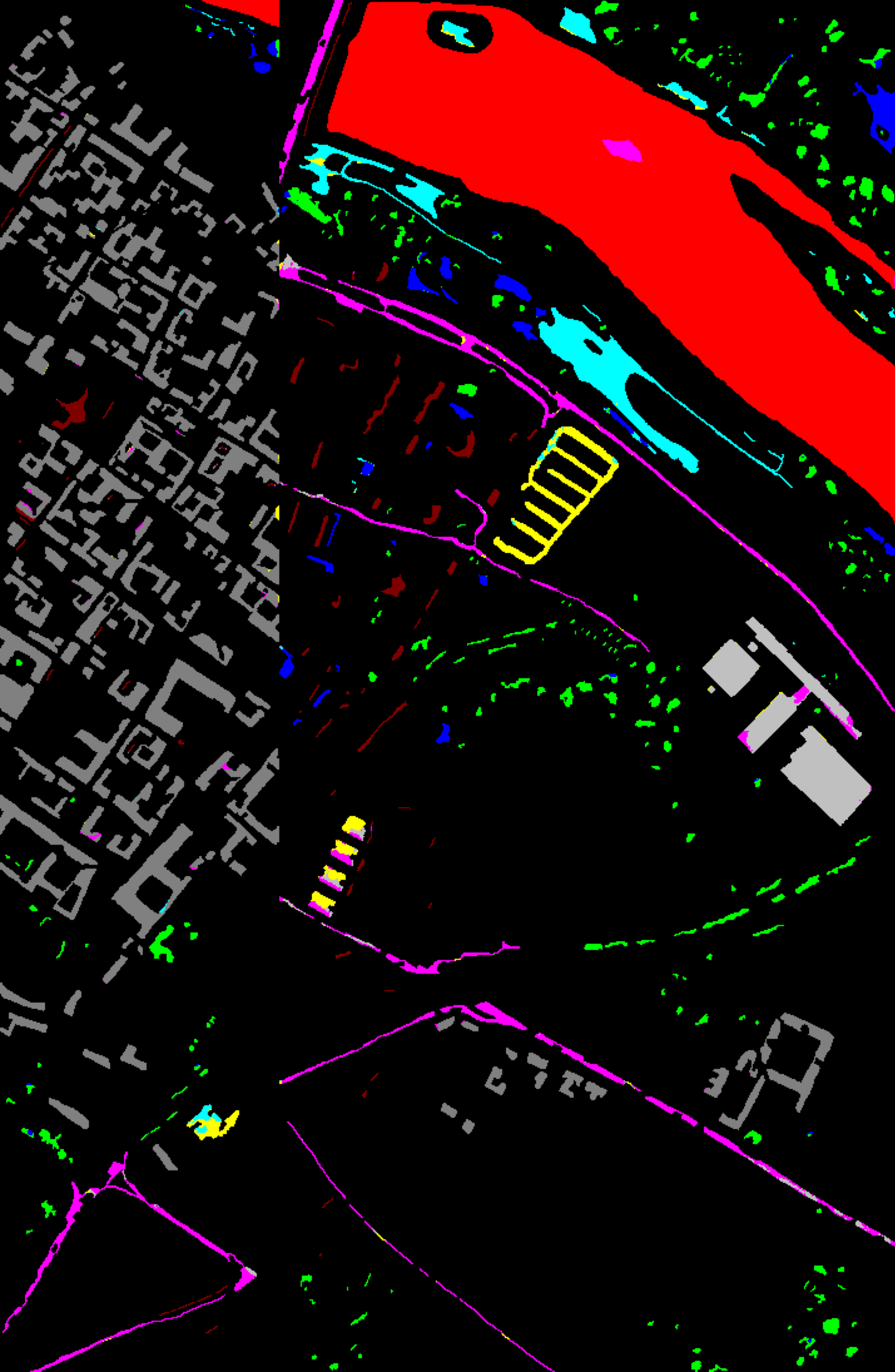}\\[0.2em]
            \tiny(h)
        \end{minipage}\hspace{\colsep}
        \begin{minipage}[t]{\colw}\centering
            \includegraphics[width=\linewidth]{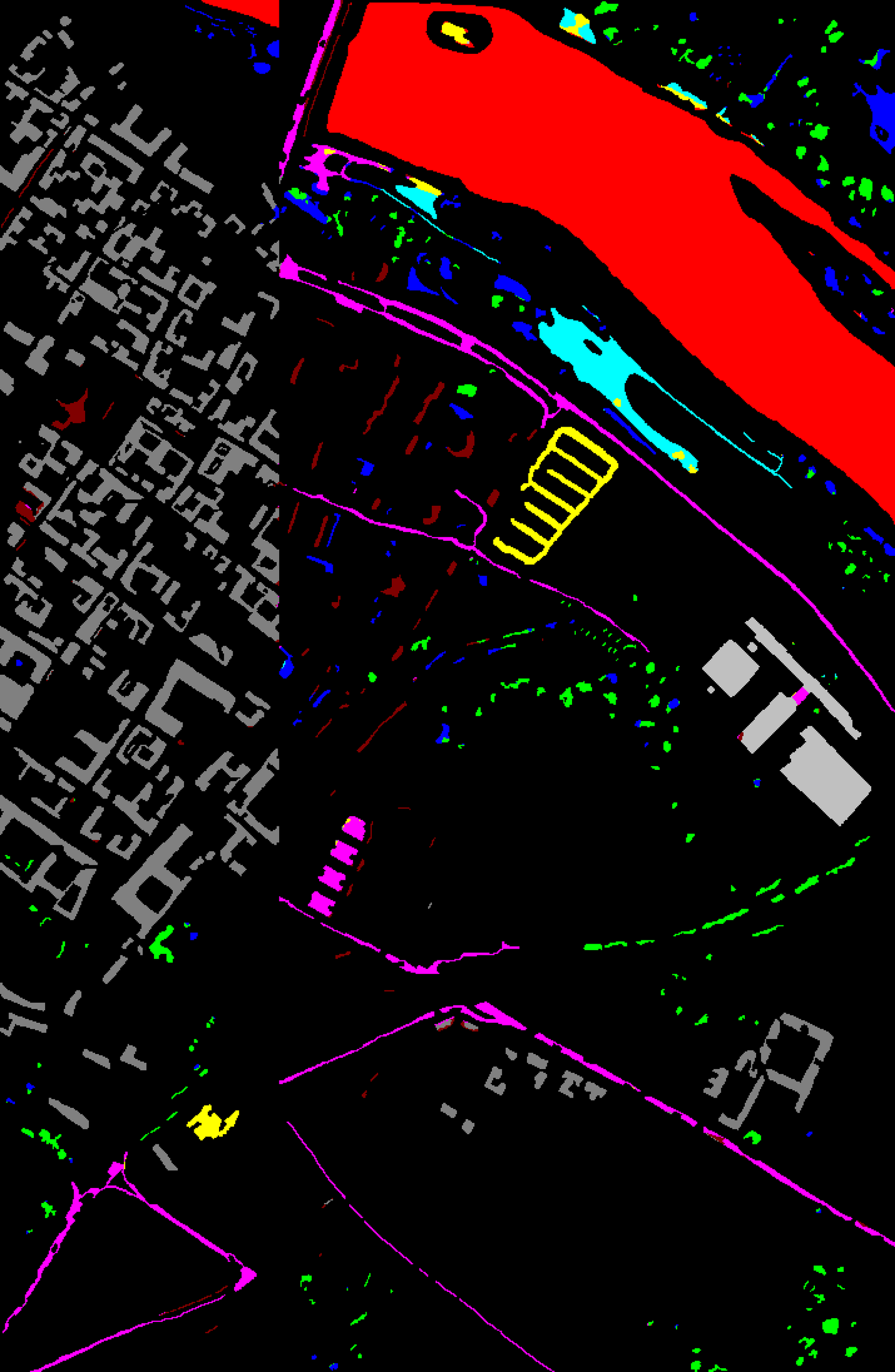}\\[0.2em]
            \tiny(i)
        \end{minipage}\hspace{\colsep}
        \begin{minipage}[t]{0.15\textwidth}\centering
            \vspace{-11em}
            \tiny
            \renewcommand{\arraystretch}{1.8}
            \begin{tabular}{@{}ll@{}}
                \textcolor[rgb]{0.0,0.0,0.0}{\rule{3mm}{3mm}}   & \textbf{Unlabelled} \\ 
                \textcolor[rgb]{1.0,0.0,0.0}{\rule{3mm}{3mm}}     & \textbf{Water} \\ 
                \textcolor[rgb]{0.0,1.0,0.0}{\rule{3mm}{3mm}}   & \textbf{Trees} \\ 
                \textcolor[rgb]{0.0,0.0,1.0}{\rule{3mm}{3mm}}& \textbf{Asphalt} \\ 
                \textcolor[rgb]{1.0,1.0,0.0}{\rule{3mm}{3mm}}  & \textbf{Bitumen} \\ 
                \textcolor[rgb]{0.0,1.0,1.0}{\rule{3mm}{3mm}}    & \textbf{Bricks} \\ 
                \textcolor[rgb]{1.0,0.0,1.0}{\rule{3mm}{3mm}} & \textbf{Tiles} \\ 
                \textcolor[rgb]{0.75,0.75,0.75}{\rule{3mm}{3mm}} & \textbf{Shadows} \\ 
                \textcolor[rgb]{0.5,0.5,0.5}{\rule{3mm}{3mm}}    & \textbf{Bare soil} \\ 
                \textcolor[rgb]{0.5,0.0,0.0}{\rule{3mm}{3mm}}   & \textbf{Meadows} \\
            \end{tabular}\\[0.8em]
            \tiny(j)
        \end{minipage}
        
    \end{minipage}
    \caption{Classification maps of different methods on PU\(\rightarrow\)PC task. (a) False color image. (b) Ground-truth. (c) SSTN. (d) CTF. (e) DEMAE. (f) DCFSL. (g) FDFSL. (h) HyMuT. (i) Ours. (j) Color labels}
    \label{fig:PC_classification_maps}

\end{figure}

1) For the PU\(\rightarrow\)PC task (Table \ref{tab:PC}), our method achieves superior comprehensive performance with exceptional global recognition and local consistency. Quantitative results show that OA achieves 96.98\% (+0.31\% over DCFSL, +0.53\% over DEMAE), AA reaches 93.30\% (+0.71\% over DEMAE, +1.92\% over CTF), Kappa equals 95.73\% (+0.73\% over DEMAE, +0.43\% DCFSL). Visually, it produces more coherent boundaries in complex regions (Figure \ref{fig:PC_classification_maps}), especially in urban-vegetation transitions. 

\begin{figure}[ht]

    \newcommand{\colw}{0.3\textwidth}
    \newcommand{\colsep}{-0.25\textwidth}
    \raggedright
    \hspace*{-11.5em}
    \begin{minipage}{2\textwidth}
        \raggedright
        \begin{minipage}[t]{\colw}\centering
            \rotatebox{-90}{\includegraphics[width=\linewidth]{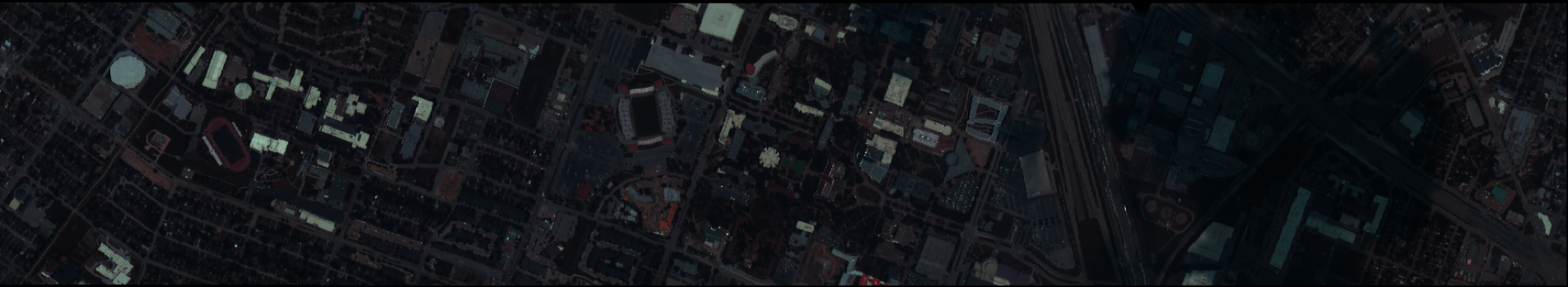}}\\[0.2em]
            \tiny(a)
        \end{minipage}\hspace{\colsep}
        \begin{minipage}[t]{\colw}\centering
            \rotatebox{-90}{\includegraphics[width=\linewidth]{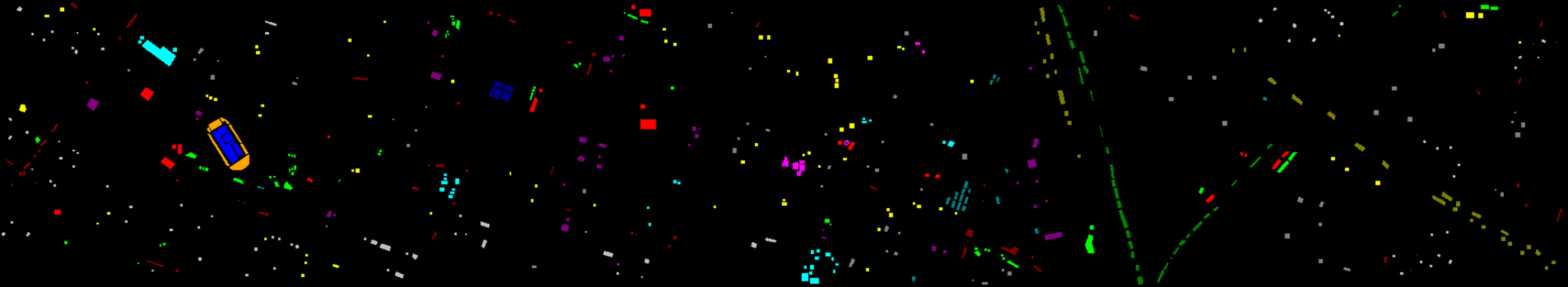}}\\[0.2em]
            \tiny(b)
        \end{minipage}\hspace{\colsep}
        \begin{minipage}[t]{\colw}\centering
            \rotatebox{-90}{\includegraphics[width=\linewidth]{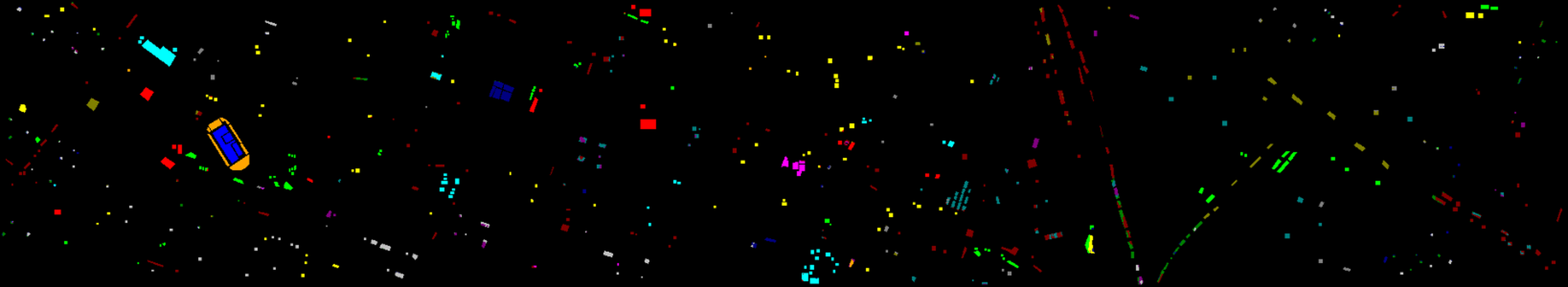}}\\[0.2em]
            \tiny(c)
        \end{minipage}\hspace{\colsep}
        \begin{minipage}[t]{\colw}\centering
            \rotatebox{-90}{\includegraphics[width=\linewidth]{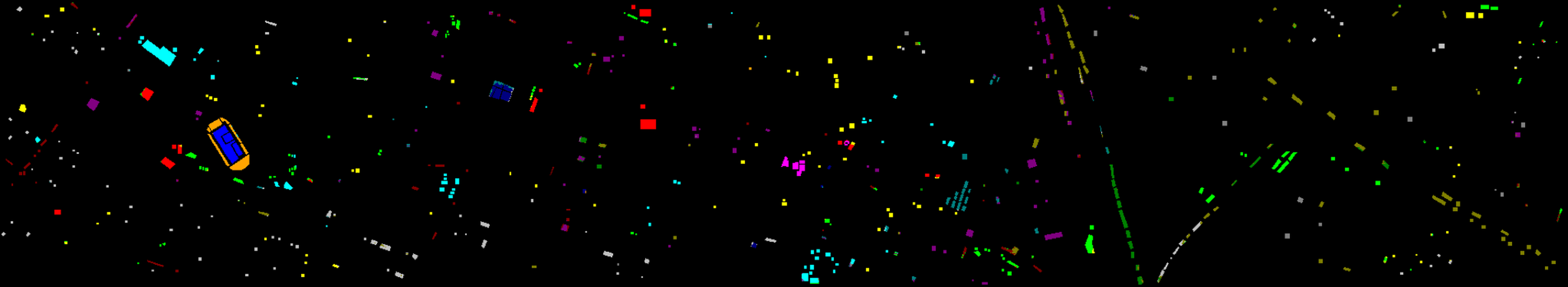}}\\[0.2em]
            \tiny(d)
        \end{minipage}\hspace{\colsep}
        \begin{minipage}[t]{\colw}\centering
            \rotatebox{-90}{\includegraphics[width=\linewidth]{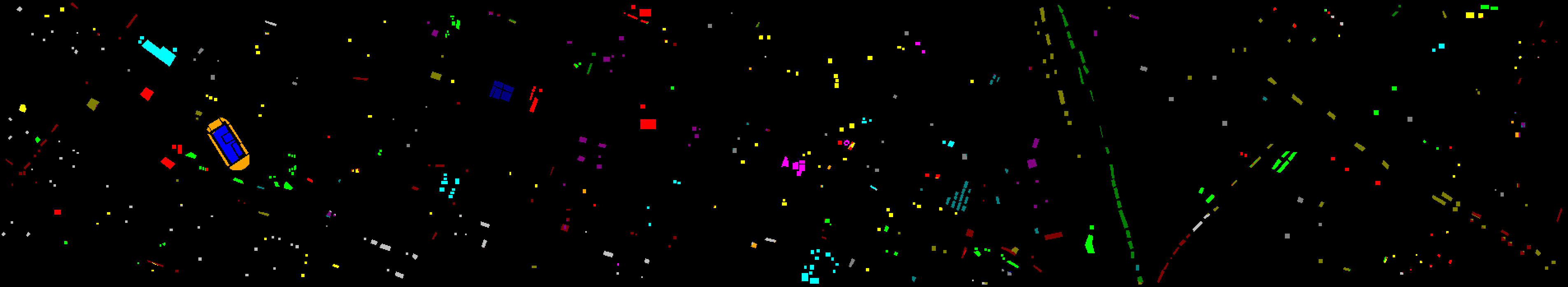}}\\[0.2em]
            \tiny(e)
        \end{minipage}\hspace{\colsep}
        \begin{minipage}[t]{\colw}\centering
            \rotatebox{-90}{\includegraphics[width=\linewidth]{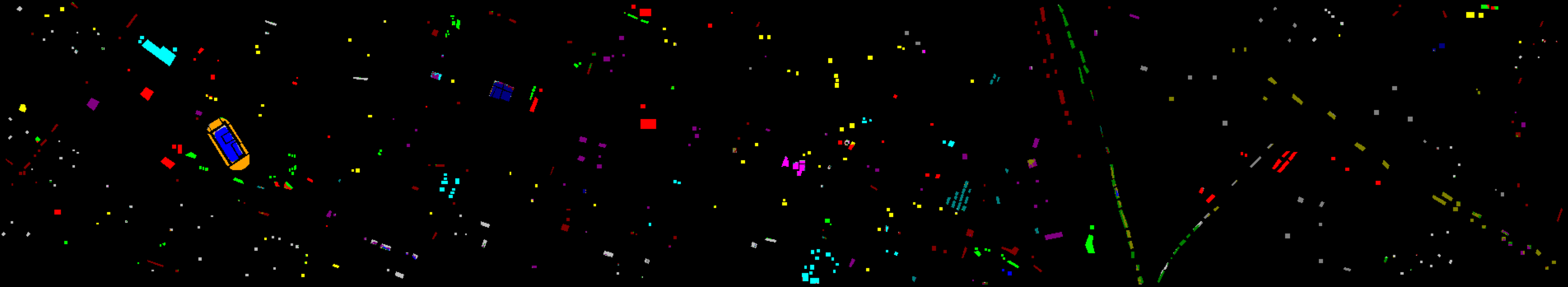}}\\[0.2em]
            \tiny(f)
        \end{minipage}\hspace{\colsep}
        \begin{minipage}[t]{\colw}\centering
            \rotatebox{-90}{\includegraphics[width=\linewidth]{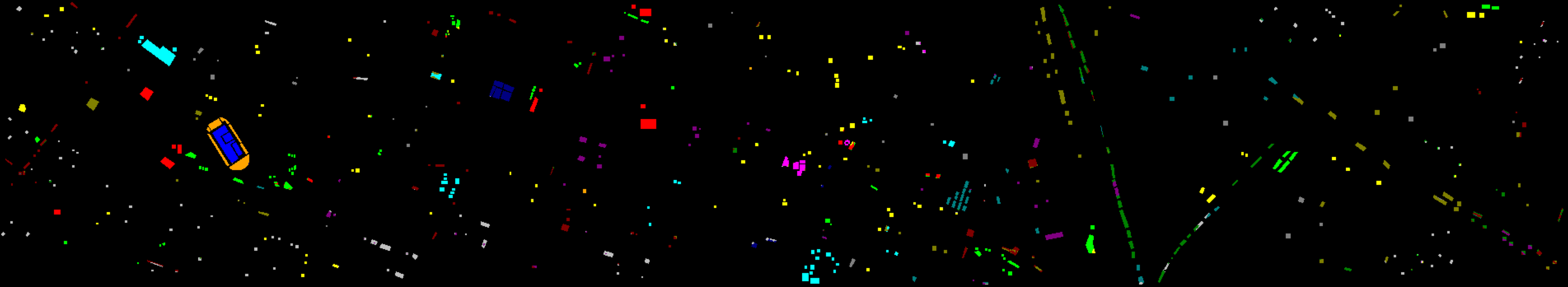}}\\[0.2em]
            \tiny(g)
        \end{minipage}\hspace{\colsep}
        \begin{minipage}[t]{\colw}\centering
            \rotatebox{-90}{\includegraphics[width=\linewidth]{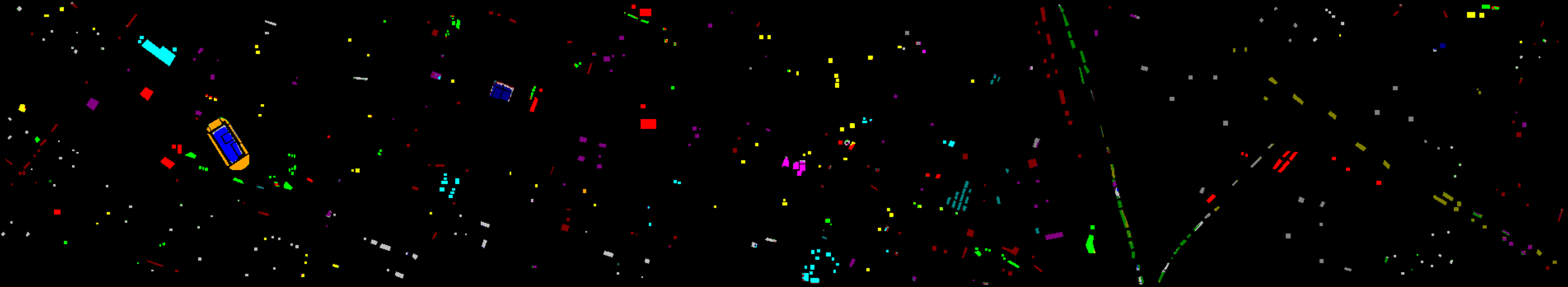}}\\[0.2em]
            \tiny(h)
        \end{minipage}\hspace{\colsep}
        \begin{minipage}[t]{\colw}\centering
            \rotatebox{-90}{\includegraphics[width=\linewidth]{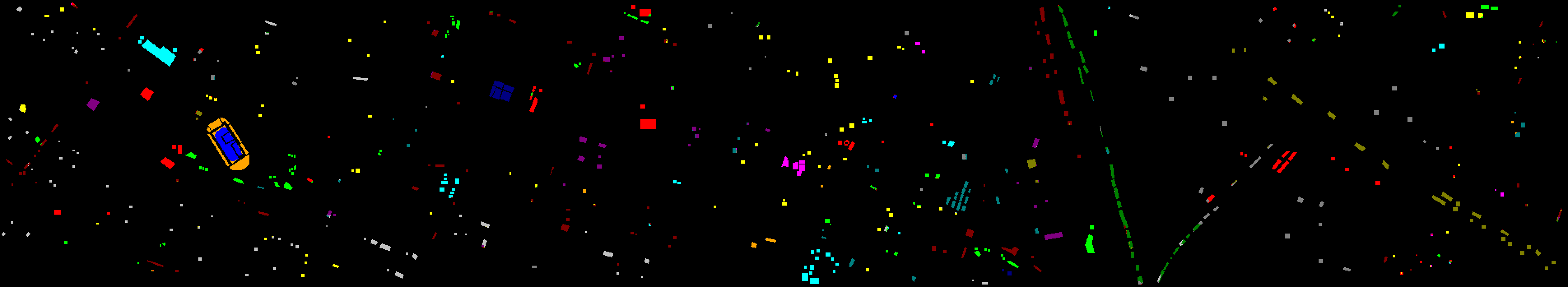}}\\[0.2em]
            \tiny(i)
        \end{minipage}\hspace{\colsep}
        \begin{minipage}[t]{0.36\textwidth}\centering
            \vspace{0.3em}
            \tiny
            \renewcommand{\arraystretch}{2}
            \begin{tabular}{@{}ll@{}}
                \textcolor[rgb]{0.0,0.0,0.0}{\rule{3mm}{3mm}}   & \textbf{Unlabelled} \\ 
                \textcolor[rgb]{1.0,0.0,0.0}{\rule{3mm}{3mm}}     & \textbf{Health grass} \\ 
                \textcolor[rgb]{0.0,1.0,0.0}{\rule{3mm}{3mm}}   & \textbf{Stressed grass} \\ 
                \textcolor[rgb]{0.0,0.0,1.0}{\rule{3mm}{3mm}}& \textbf{Synthetic grass} \\ 
                \textcolor[rgb]{1.0,1.0,0.0}{\rule{3mm}{3mm}}  & \textbf{Trees} \\ 
                \textcolor[rgb]{0.0,1.0,1.0}{\rule{3mm}{3mm}}    & \textbf{Soil} \\ 
                \textcolor[rgb]{1.0,0.0,1.0}{\rule{3mm}{3mm}} & \textbf{Water} \\ 
                \textcolor[rgb]{0.75,0.75,0.75}{\rule{3mm}{3mm}} & \textbf{Residential} \\ 
                \textcolor[rgb]{0.5,0.5,0.5}{\rule{3mm}{3mm}}    & \textbf{Commercial} \\ 
                \textcolor[rgb]{0.5,0.0,0.0}{\rule{3mm}{3mm}}   & \textbf{Road} \\
                \textcolor[rgb]{0.5,0.5,0.0}{\rule{3mm}{3mm}}   & \textbf{Highway} \\
                \textcolor[rgb]{0.0,0.5,0.0}{\rule{3mm}{3mm}}   & \textbf{Railway} \\
                \textcolor[rgb]{0.5,0.0,0.5}{\rule{3mm}{3mm}}   & \textbf{Parking lot1} \\
                \textcolor[rgb]{0.0,0.5,0.5}{\rule{3mm}{3mm}}   & \textbf{Parking lot2} \\
                \textcolor[rgb]{0.0,0.0,0.5}{\rule{3mm}{3mm}}   & \textbf{Tennis court} \\
                \textcolor[rgb]{1.0,0.65,0.0}{\rule{3mm}{3mm}}   & \textbf{Running track} \\
            \end{tabular}\\[1.0em]
            \tiny(j)
        \end{minipage}
        
    \end{minipage}
    \caption{Classification maps of different methods on the SA\(\rightarrow\)HU task. (a) False color image. (b) Ground-truth. (c) SSTN. (d) CTF. (e) DEMAE. (f) DCFSL. (g) FDFSL. (h) HyMuT. (i) Ours. (j) Color labels.}
    \label{fig:HU_classification_maps}
    
\end{figure}

2) For the chellenging SA\(\rightarrow\)HU task with high spectral similarity and minimal material variation (Table \ref{tab:HU}), our model attains 79.88\% OA, 82.89\% AA, and 78.28\% Kappa, gaining  average approximately 2\%–3\% average improvement over DEMAE and CTF. It exhibits enhanced robustness in low-contrast regions, producing smoother boundaries and better structural detail preservation (Figure \ref{fig:HU_classification_maps}).

\begin{figure}[htbp]
    
    \newcommand{\colw}{0.18\textwidth}
    \newcommand{\colsep}{-0.005\textwidth}
    \raggedright
    \hspace*{-2em}
    \begin{minipage}{1.2\textwidth}

        \begin{minipage}[t]{\colw}\centering
            \includegraphics[width=\linewidth]{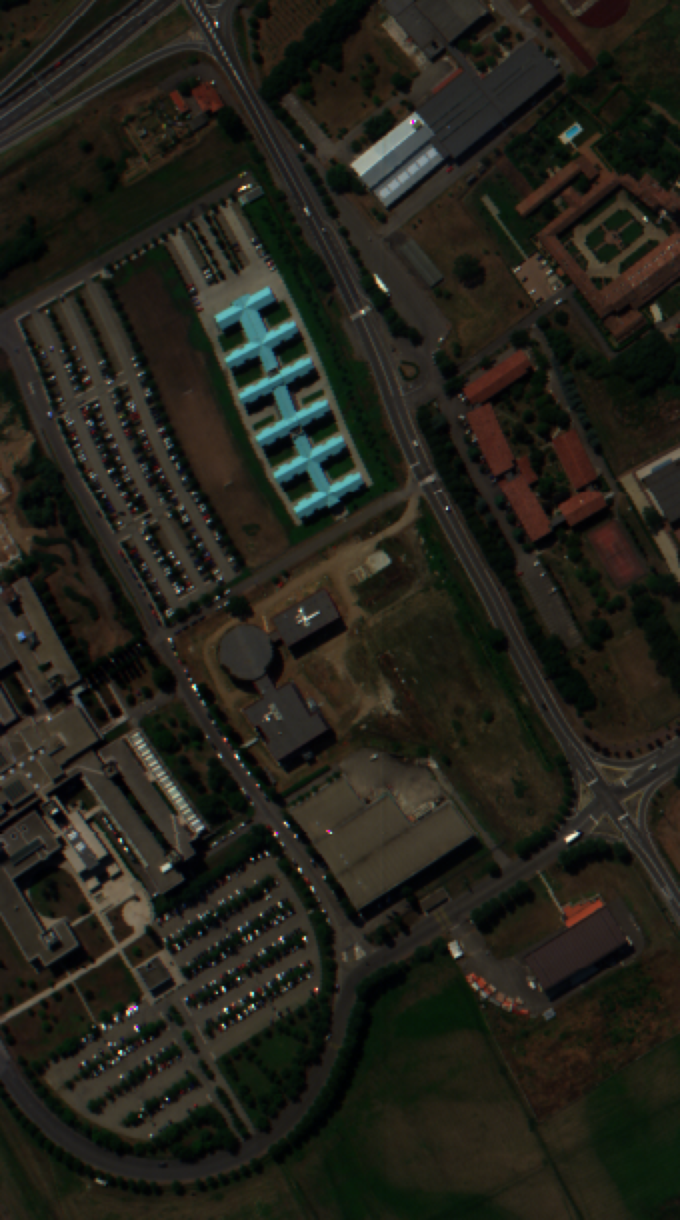}\\[0.2em]
            \tiny(a)
        \end{minipage}\hspace{\colsep}
        \begin{minipage}[t]{\colw}\centering
            \includegraphics[width=\linewidth]{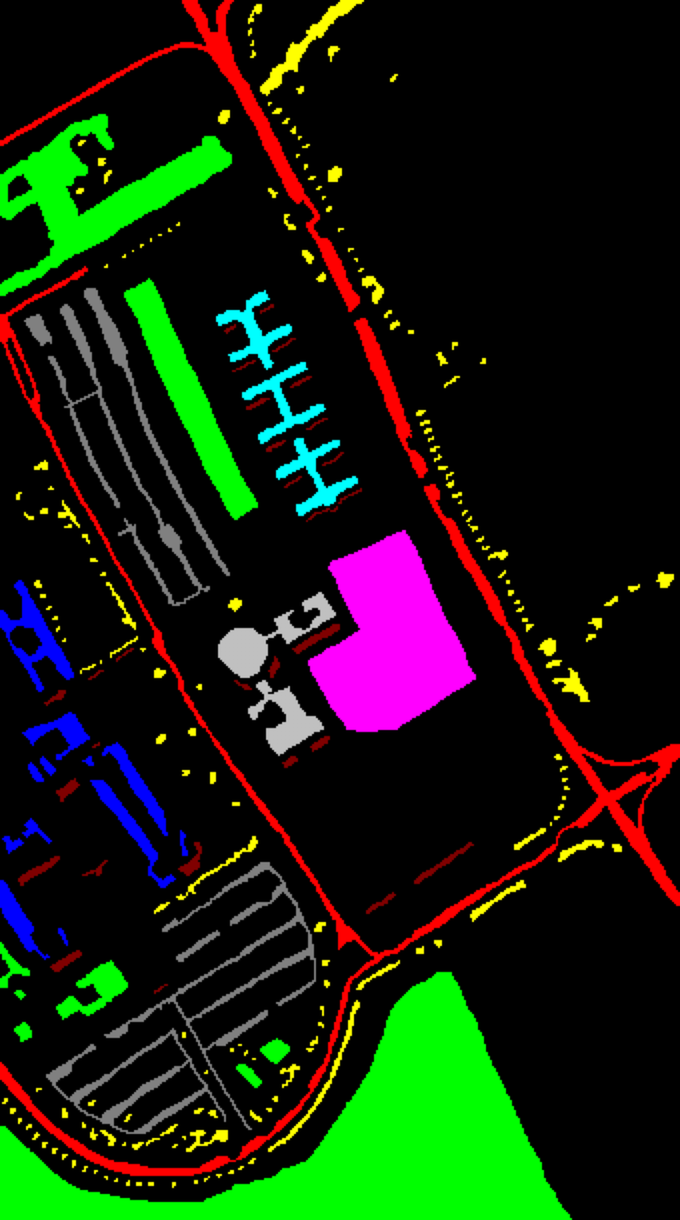}\\[0.2em]
            \tiny(b)
        \end{minipage}\hspace{\colsep}
        \begin{minipage}[t]{\colw}\centering
            \includegraphics[width=\linewidth]{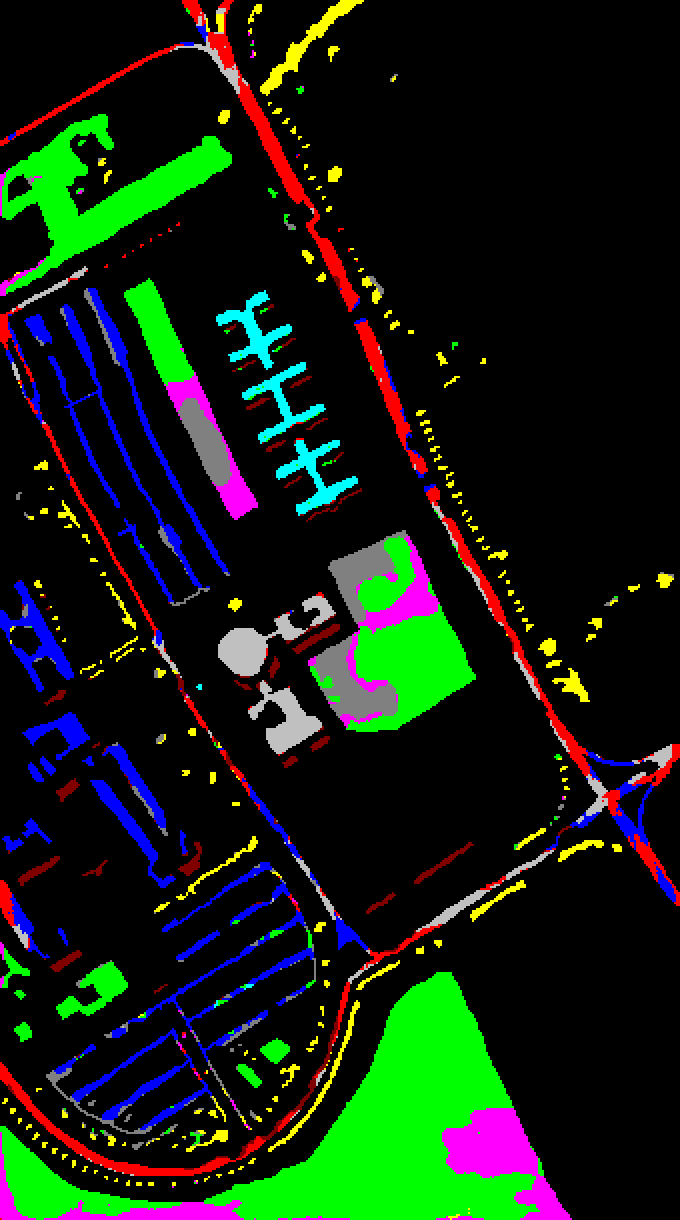}\\[0.2em]
            \tiny(c)
        \end{minipage}\hspace{\colsep}
        \begin{minipage}[t]{\colw}\centering
            \includegraphics[width=\linewidth]{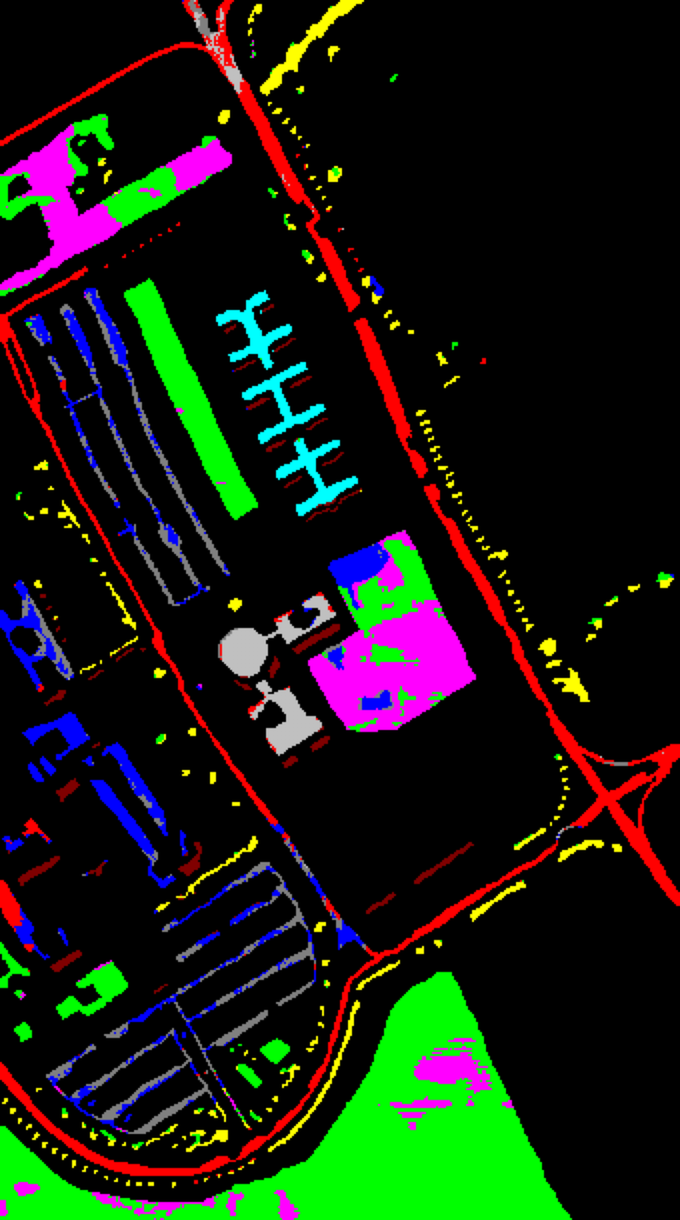}\\[0.2em]
            \tiny(d)
        \end{minipage}\hspace{\colsep}
        \begin{minipage}[t]{\colw}\centering
            \includegraphics[width=\linewidth]{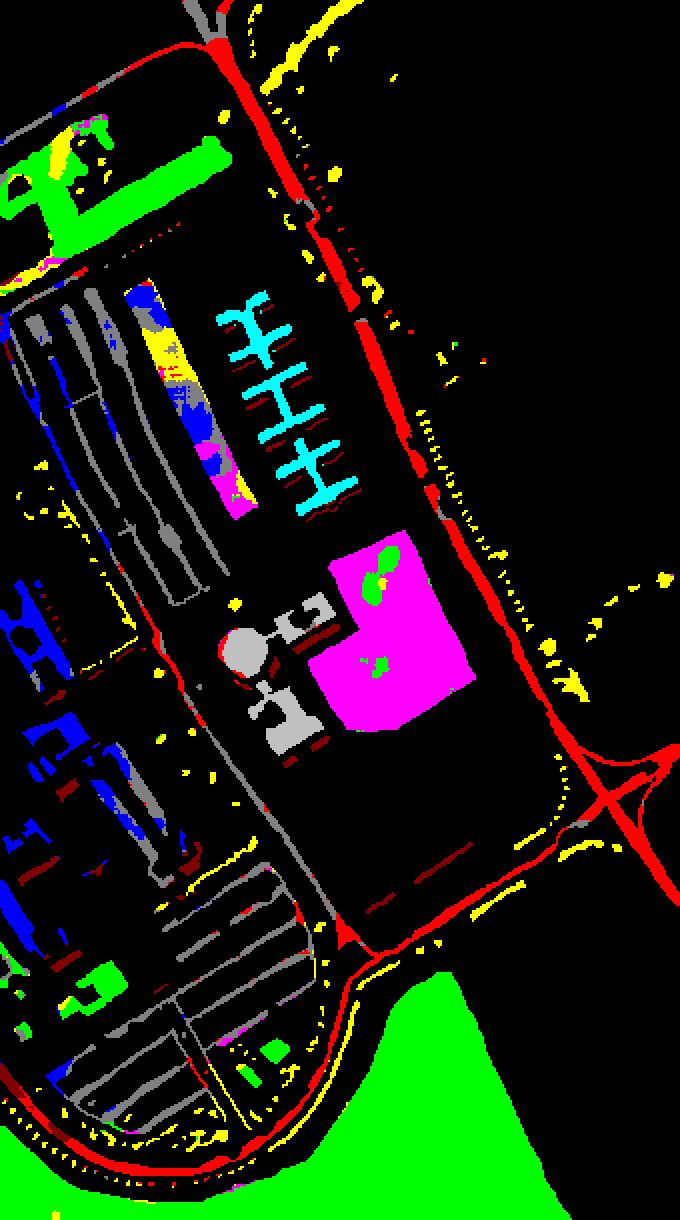}\\[0.2em]
            \tiny(e)
        \end{minipage}
    
        \vspace{0.8em}
    
        \begin{minipage}[t]{\colw}\centering
            \includegraphics[width=\linewidth]{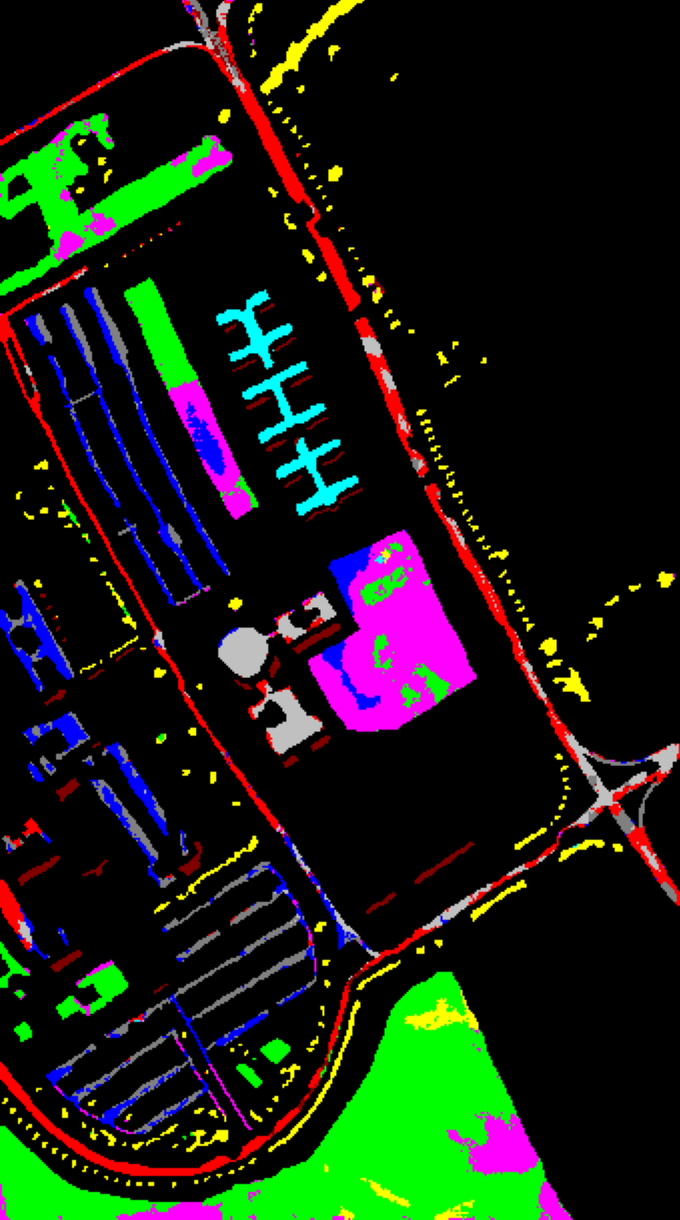}\\[0.2em]
            \tiny(f)
        \end{minipage}\hspace{\colsep}
        \begin{minipage}[t]{\colw}\centering
            \includegraphics[width=\linewidth]{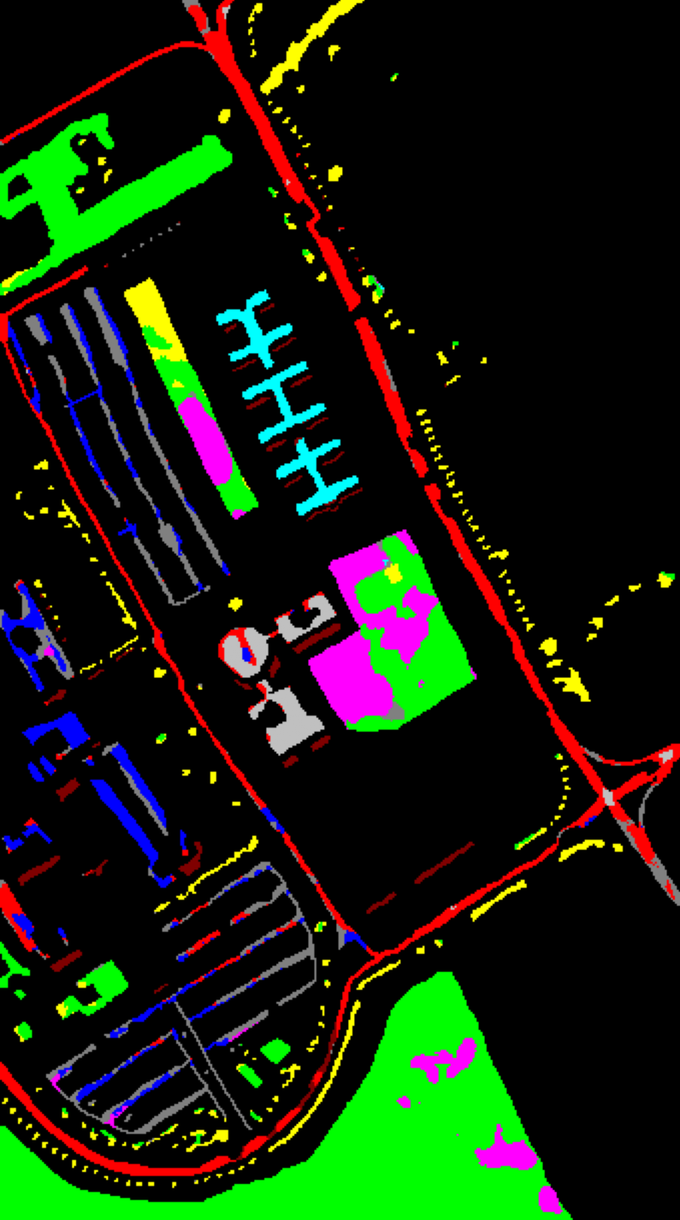}\\[0.2em]
            \tiny(g)
        \end{minipage}\hspace{\colsep}
        \begin{minipage}[t]{\colw}\centering
            \includegraphics[width=\linewidth]{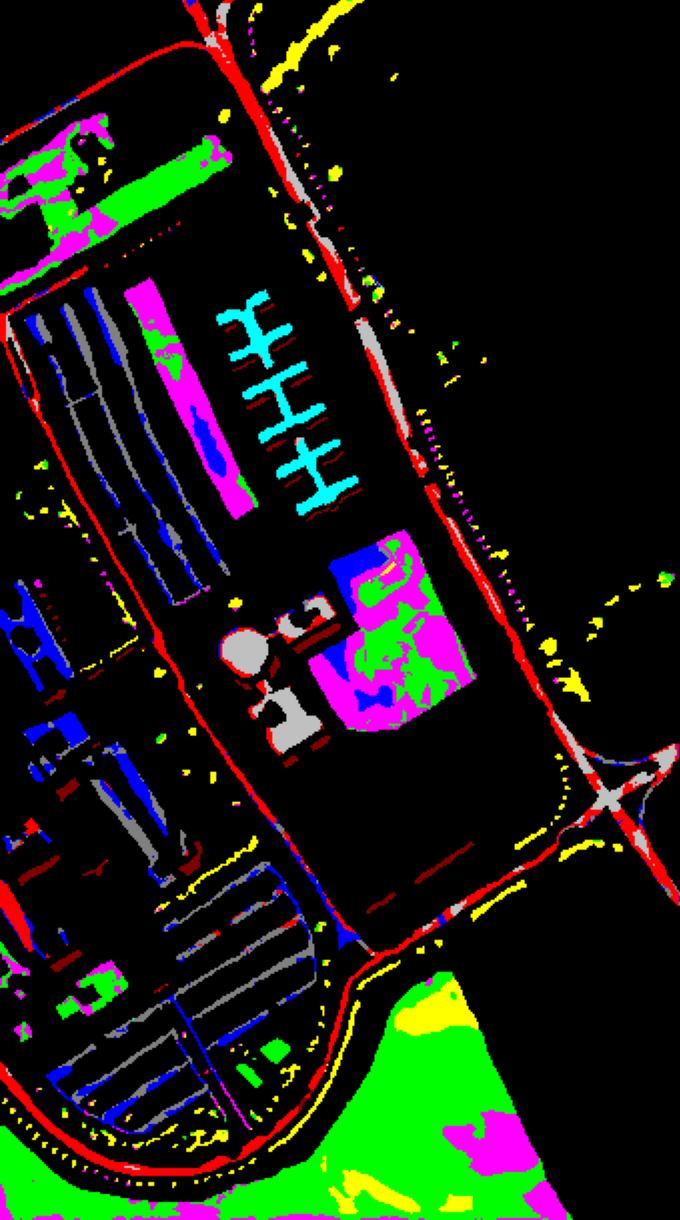}\\[0.2em]
            \tiny(h)
        \end{minipage}\hspace{\colsep}
        \begin{minipage}[t]{\colw}\centering
            \includegraphics[width=\linewidth]{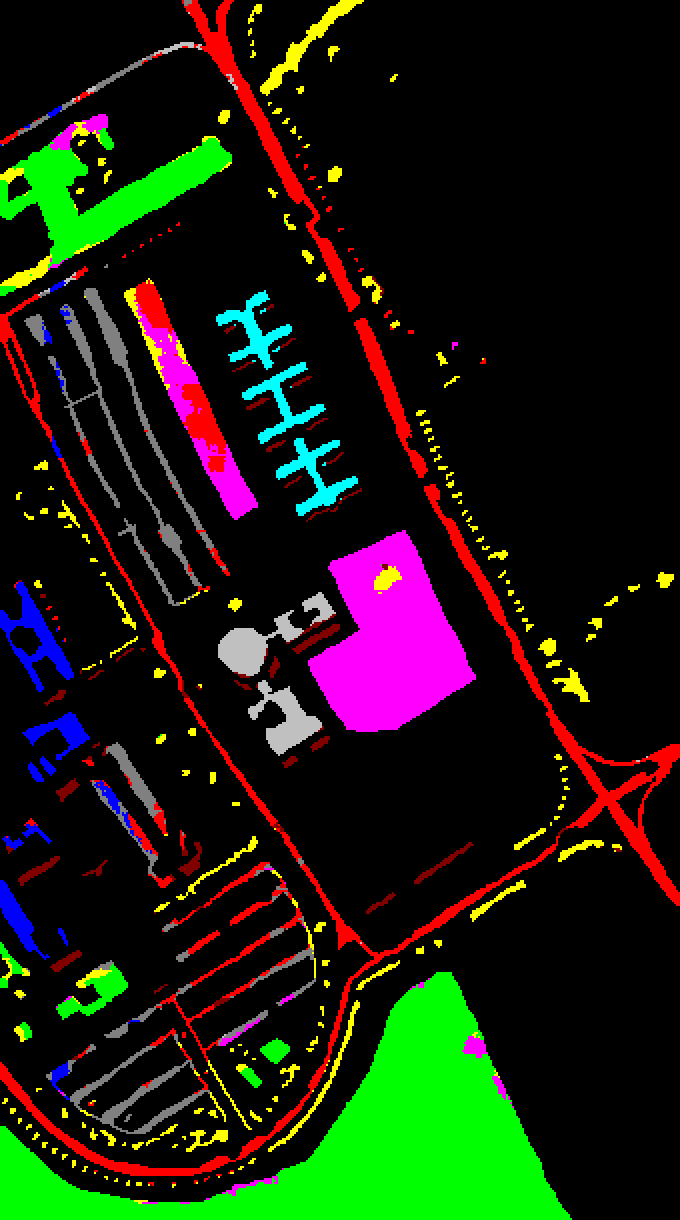}\\[0.2em]
            \tiny(i)
        \end{minipage}\hspace{\colsep}
        \begin{minipage}[t]{0.17\textwidth}\centering
            \vspace{-12.5em}
            \tiny
            \renewcommand{\arraystretch}{2.1}
            \begin{tabular}{@{}ll@{}}
                \textcolor[rgb]{0.0,0.0,0.0}{\rule{3mm}{3mm}}   & \textbf{Unlabelled} \\ 
                \textcolor[rgb]{1.0,0.0,0.0}{\rule{3mm}{3mm}}     & \textbf{Asphalt} \\ 
                \textcolor[rgb]{0.0,1.0,0.0}{\rule{3mm}{3mm}}   & \textbf{Meadows} \\ 
                \textcolor[rgb]{0.0,0.0,1.0}{\rule{3mm}{3mm}}& \textbf{Gravel} \\ 
                \textcolor[rgb]{1.0,1.0,0.0}{\rule{3mm}{3mm}}  & \textbf{Trees} \\ 
                \textcolor[rgb]{0.0,1.0,1.0}{\rule{3mm}{3mm}}    & \textbf{Metal} \\ 
                \textcolor[rgb]{1.0,0.0,1.0}{\rule{3mm}{3mm}} & \textbf{Bare Soil} \\ 
                \textcolor[rgb]{0.75,0.75,0.75}{\rule{3mm}{3mm}} & \textbf{Bitumen} \\ 
                \textcolor[rgb]{0.5,0.5,0.5}{\rule{3mm}{3mm}}    & \textbf{Bricks} \\ 
                \textcolor[rgb]{0.5,0.0,0.0}{\rule{3mm}{3mm}}   & \textbf{Shadows} \\
            \end{tabular}\\[0.5em]
            \tiny(j)
        \end{minipage}
        
    \end{minipage}
    \caption{Classification maps of different methods on HU\(\rightarrow\)PU task. (a) False color image. (b) Ground-truth. (c) SSTN. (d) CTF. (e) DEMAE. (f) DCFSL. (g) FDFSL. (h) HyMuT. (i) Ours. (j) Color labels}
    \label{fig:PU_classification_maps}
    
\end{figure}

3) For the HU\(\rightarrow\)PU task presented in Table~\ref{tab:PU}, our model delivers robust performance with 84.82\% OA, 87.03\% AA, and 80.44\% Kappa, outperforming DEMAE and CTF. It excels in classifying spectral-variable features (Trees: 86.23\%, Bare Soil: 99.90\%), confirming robustness to shadow and texture noise. Visual results (Figure~\ref{fig:PU_classification_maps}) show precise vegetation and bare soil delineation with natural transitions and boundary continuity.

\begin{figure}[htbp]
    
    \newcommand{\colw}{0.18\textwidth}
    \newcommand{\colsep}{-0.005\textwidth}
    \raggedright
    \hspace*{-2.8em}
    \begin{minipage}{1.2\textwidth}

        \begin{minipage}[t]{\colw}\centering
            \includegraphics[width=\linewidth]{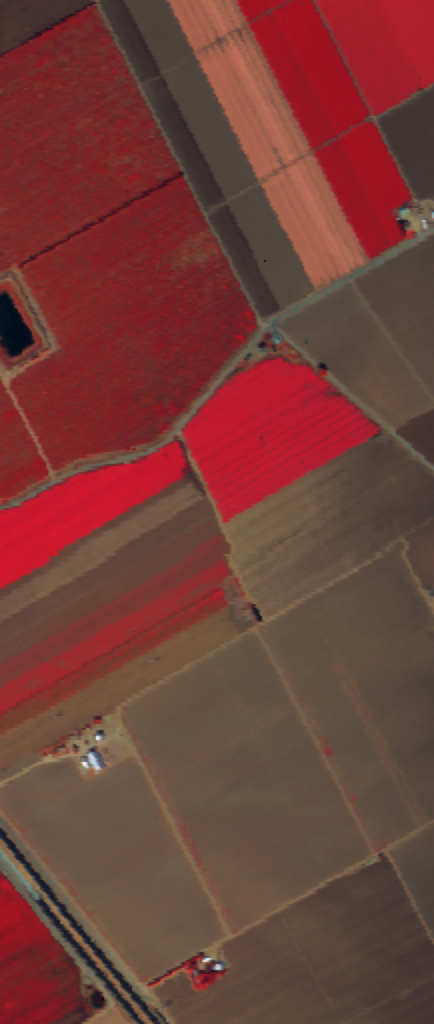}\\[0.2em]
            \tiny(a)
        \end{minipage}\hspace{\colsep}
        \begin{minipage}[t]{\colw}\centering
            \includegraphics[width=\linewidth]{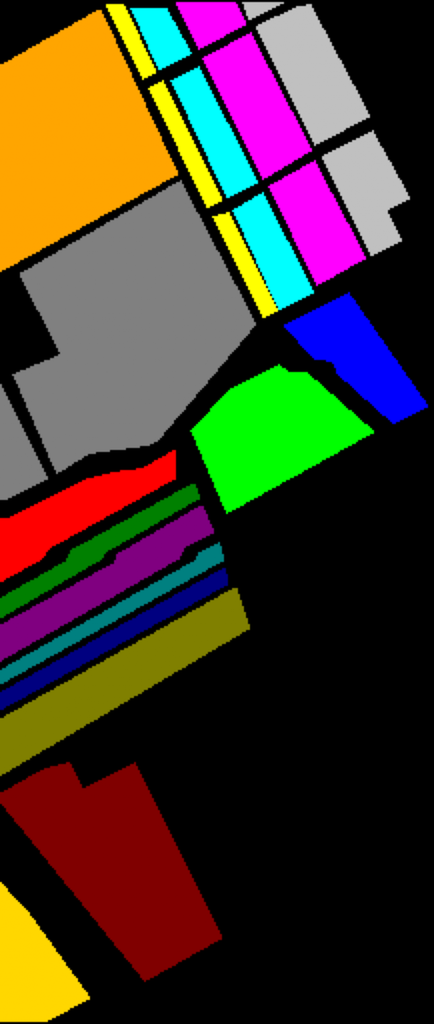}\\[0.2em]
            \tiny(b)
        \end{minipage}\hspace{\colsep}
        \begin{minipage}[t]{\colw}\centering
            \includegraphics[width=\linewidth]{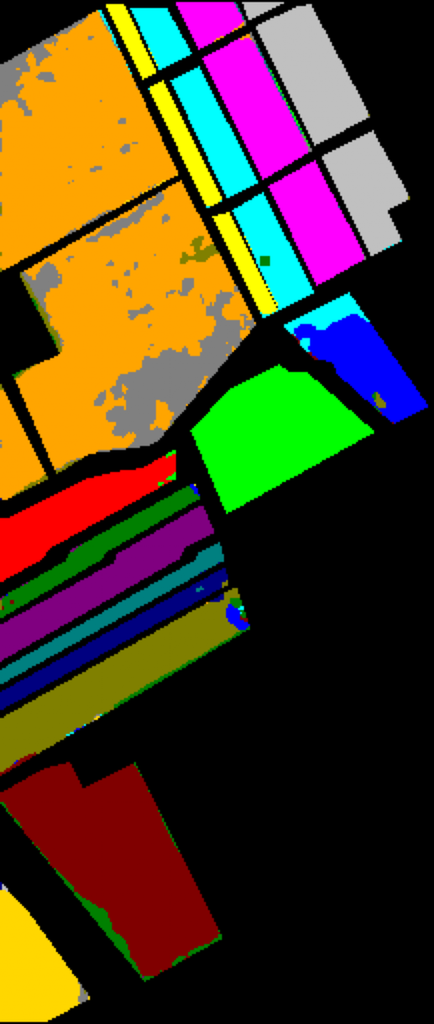}\\[0.2em]
            \tiny(c)
        \end{minipage}\hspace{\colsep}
        \begin{minipage}[t]{\colw}\centering
            \includegraphics[width=\linewidth]{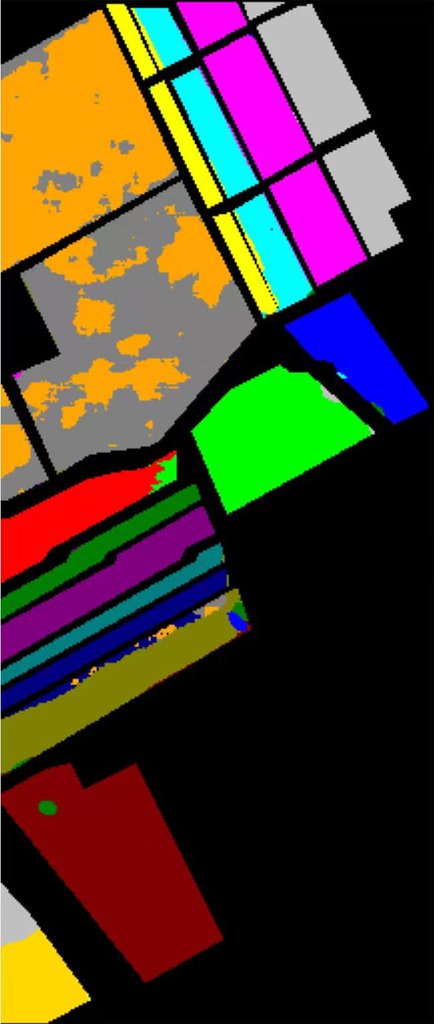}\\[0.2em]
            \tiny(d)
        \end{minipage}\hspace{\colsep}
        \begin{minipage}[t]{\colw}\centering
            \includegraphics[width=\linewidth]{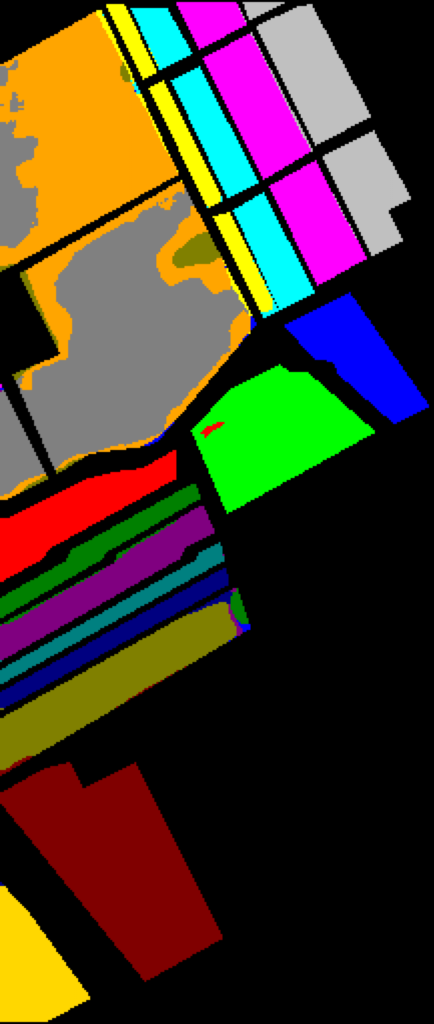}\\[0.2em]
            \tiny(e)
        \end{minipage}
    
        \vspace{0.8em}
    
        \begin{minipage}[t]{\colw}\centering
            \includegraphics[width=\linewidth]{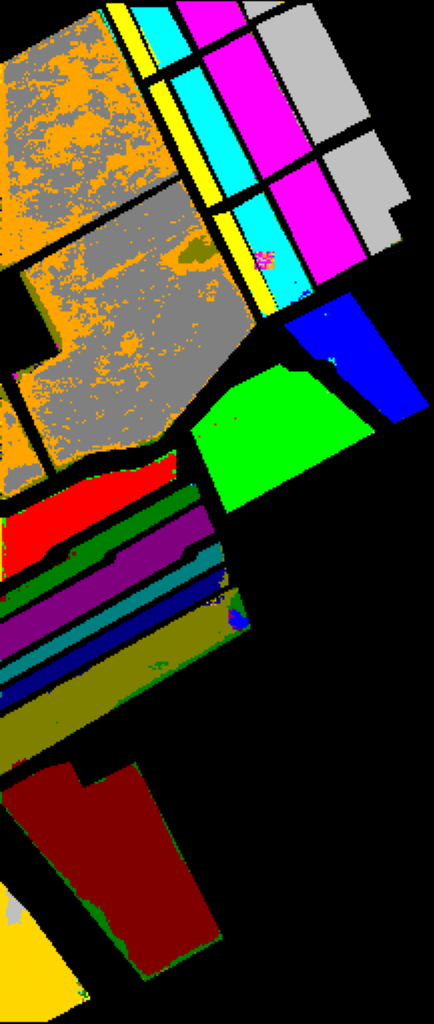}\\[0.2em]
            \tiny(f)
        \end{minipage}\hspace{\colsep}
        \begin{minipage}[t]{\colw}\centering
            \includegraphics[width=\linewidth]{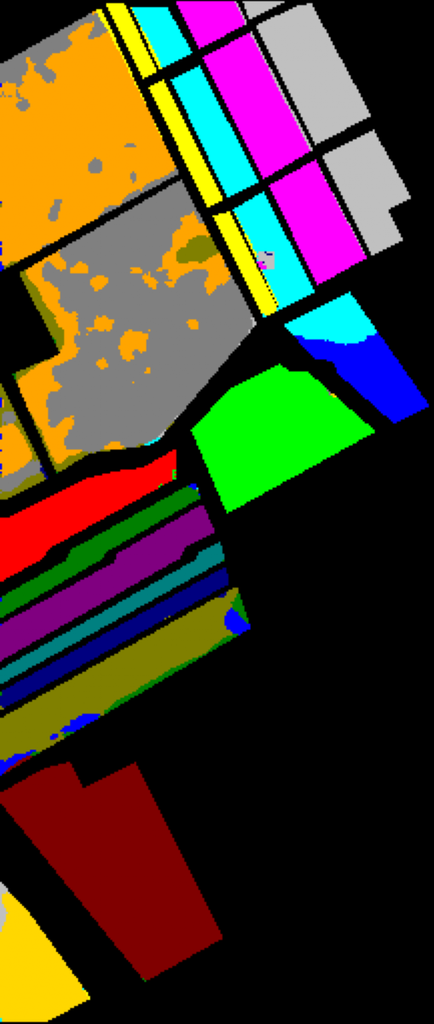}\\[0.2em]
            \tiny(g)
        \end{minipage}\hspace{\colsep}
        \begin{minipage}[t]{\colw}\centering
            \includegraphics[width=\linewidth]{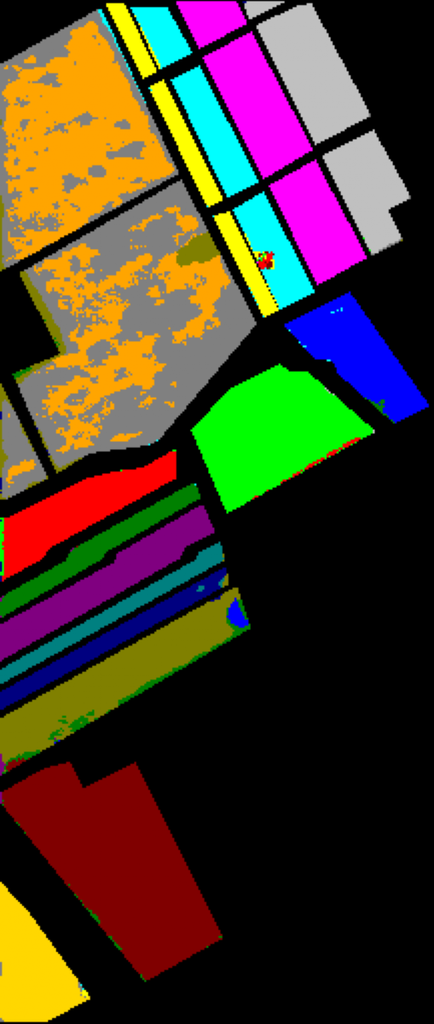}\\[0.2em]
            \tiny(h)
        \end{minipage}\hspace{\colsep}
        \begin{minipage}[t]{\colw}\centering
            \includegraphics[width=\linewidth]{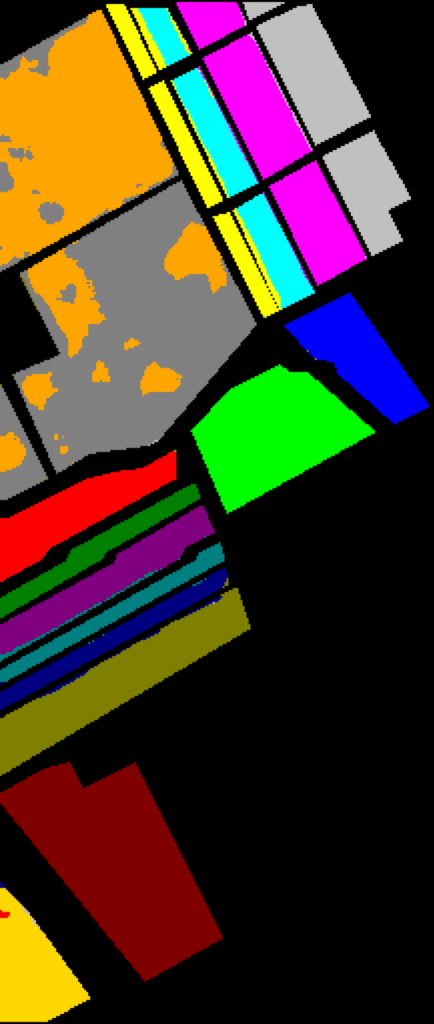}\\[0.2em]
            \tiny(i)
        \end{minipage}\hspace{\colsep}
        \begin{minipage}[t]{\colw}\centering
            \vspace{-16.5em}
            \scriptsize
            \scalebox{0.8}{
            \renewcommand{\arraystretch}{1.5} 
            \begin{tabular}{@{}ll@{}}
                \textcolor[rgb]{0.0,0.0,0.0}{\rule{3mm}{3mm}}   & \textbf{Unlabelled} \\ 
                \textcolor[rgb]{1.0,0.0,0.0}{\rule{3mm}{3mm}}   & \textbf{Brocoli green weeds 1} \\ 
                \textcolor[rgb]{0.0,1.0,0.0}{\rule{3mm}{3mm}}   & \textbf{Brocoli green weeds 2} \\ 
                \textcolor[rgb]{0.0,0.0,1.0}{\rule{3mm}{3mm}}   & \textbf{Fallow} \\ 
                \textcolor[rgb]{1.0,1.0,0.0}{\rule{3mm}{3mm}}   & \textbf{Fallow rough plow} \\ 
                \textcolor[rgb]{0.0,1.0,1.0}{\rule{3mm}{3mm}}   & \textbf{Fallow smooth} \\ 
                \textcolor[rgb]{1.0,0.0,1.0}{\rule{3mm}{3mm}}   & \textbf{Stubble} \\ 
                \textcolor[rgb]{0.75,0.75,0.75}{\rule{3mm}{3mm}} & \textbf{Celery} \\ 
                \textcolor[rgb]{0.5,0.5,0.5}{\rule{3mm}{3mm}}   & \textbf{Grapes untarined} \\ 
                \textcolor[rgb]{0.5,0.0,0.0}{\rule{3mm}{3mm}}   & \textbf{Soil vineyard develop} \\ 
                \textcolor[rgb]{0.5,0.5,0.0}{\rule{3mm}{3mm}}   & \textbf{Corn senesced green weeds} \\ 
                \textcolor[rgb]{0.0,0.5,0.0}{\rule{3mm}{3mm}}   & \textbf{Lettuce romaine 4wk} \\ 
                \textcolor[rgb]{0.5,0.0,0.5}{\rule{3mm}{3mm}}   & \textbf{Lettuce romaine 5wk} \\ 
                \textcolor[rgb]{0.0,0.5,0.5}{\rule{3mm}{3mm}}   & \textbf{Lettuce romaine 6wk} \\ 
                \textcolor[rgb]{0.0,0.0,0.5}{\rule{3mm}{3mm}}   & \textbf{Lettuce romaine 7wk} \\ 
                \textcolor[rgb]{1.0,0.65,0.0}{\rule{3mm}{3mm}}  & \textbf{Vineyard untrained} \\
                \textcolor[rgb]{1.0,0.85,0.0}{\rule{3mm}{3mm}}   & \textbf{Vineyard vertical trellis} \\
            \end{tabular}}\\[0.5em]
            \tiny(j)
        \end{minipage}

    \end{minipage}
    \caption{Classification maps of different methods on PC\(\rightarrow\)SA task. (a) False color image. (b) Ground-truth. (c) SSTN. (d) CTF. (e) DEMAE. (f) DCFSL. (g) FDFSL. (h) HyMuT. (i) Ours. (j) Color labels}
    \label{fig:SA_classification_maps}

\end{figure}

4) For the PC\(\rightarrow\)SA task with complex land cover and uneven distribution (Table \ref{tab:SA}), our method achieves 93.05\% OA, 96.19\% AA, and 92.28\% Kappa, showing distinct advantage in discriminating spectrally similar categories. Despite DEMAE's local representation capability and SSTN's texture robustness, our approach overcomes their limitations in local fragmentation and boundary ambiguity. Visual results (Figure \ref{fig:SA_classification_maps}) confirm superior global consistency and balanced class representation. 

\begin{table}[H]
\centering
\caption{Classification results on PU training -- PC testing task}
\label{tab:PC}
\renewcommand{\arraystretch}{1.2}
\setlength{\tabcolsep}{6pt}
\resizebox{\textwidth}{!}{
\begin{tabular}{c|ccccccc}
\hline
\textbf{Metric} & \textbf{SSTN} & \textbf{CTF} & \textbf{DEMAE} & \textbf{DCFSL} & \textbf{FDFSL} & \textbf{HyMuT} & \textbf{Ours} \\
\hline
1 & 95.76 $\pm$ 1.73 & 98.11 $\pm$ 1.33 & 99.50 $\pm$ 0.57 & 99.63 $\pm$ 0.17 & 98.76 $\pm$ 0.59 & 99.63 $\pm$ 0.26 & \textbf{99.98 $\pm$ 0.06} \\
2 & 67.30 $\pm$ 23.64 & 85.25 $\pm$ 12.23 & 88.12 $\pm$ 4.98 & 90.25 $\pm$ 5.56 & 81.24 $\pm$ 4.74 & \textbf{90.35 $\pm$ 3.94} & 84.04 $\pm$ 7.24 \\
3 & 94.17 $\pm$ 3.90 & \textbf{94.81 $\pm$ 1.84} & 85.82 $\pm$ 17.65 & 91.12 $\pm$ 2.19 & 80.19 $\pm$ 9.03 & 91.05 $\pm$ 5.77 & 94.71 $\pm$ 3.33 \\
4 & 63.61 $\pm$ 16.52 & 86.49 $\pm$ 3.34 & \textbf{99.50 $\pm$ 0.71} & 96.23 $\pm$ 2.78 & 79.42 $\pm$ 18.66 & 89.40 $\pm$ 8.73 & 97.25 $\pm$ 2.96 \\
5 & \textbf{91.26 $\pm$ 1.31} & 90.42 $\pm$ 5.47 & 82.66 $\pm$ 8.75 & 85.06 $\pm$ 3.27 & 76.07 $\pm$ 8.34 & 90.39 $\pm$ 5.82 & 87.52 $\pm$ 7.63 \\
6 & \textbf{97.18 $\pm$ 1.66} & 95.65 $\pm$ 2.55 & 95.26 $\pm$ 4.16 & 95.66 $\pm$ 1.50 & 83.19 $\pm$ 8.90 & 90.43 $\pm$ 9.62 & 91.82 $\pm$ 8.86 \\
7 & 90.95 $\pm$ 2.69 & 81.33 $\pm$ 4.68 & 88.99 $\pm$ 5.40 & 84.68 $\pm$ 2.34 & 79.76 $\pm$ 4.66 & 83.03 $\pm$ 1.94 & \textbf{91.99 $\pm$ 6.16} \\
8 & 97.27 $\pm$ 1.35 & 94.42 $\pm$ 3.79 & 97.48 $\pm$ 2.80 & 97.54 $\pm$ 1.31 & 86.19 $\pm$ 8.37 & 97.17 $\pm$ 1.66 & \textbf{98.41 $\pm$ 1.33} \\
9 & 96.29 $\pm$ 4.26 & 95.96 $\pm$ 1.69 & 95.92 $\pm$ 2.63 & \textbf{99.31 $\pm$ 1.06} & 92.15 $\pm$ 5.39 & 98.60 $\pm$ 0.85 & 94.00 $\pm$ 2.19 \\
\hline
OA & 93.80 $\pm$ 1.54 & 94.74 $\pm$ 1.56 & 96.45 $\pm$ 0.82 & 96.67 $\pm$ 0.61 & 90.45 $\pm$ 2.32 & 96.26 $\pm$ 0.26 & \textbf{96.98 $\pm$ 0.67} \\
AA & 88.26 $\pm$ 4.55 & 91.38 $\pm$ 2.59 & 92.59 $\pm$ 2.33 & 93.28 $\pm$ 0.55 & 84.11 $\pm$ 2.30 & 92.33 $\pm$ 1.76 & \textbf{93.30 $\pm$ 1.34} \\
$\kappa\times100$ & 91.33 $\pm$ 2.15 & 92.62 $\pm$ 2.20 & 95.00 $\pm$ 1.15 & 95.30 $\pm$ 0.85 & 86.71 $\pm$ 3.12 & 94.73 $\pm$ 1.20 & \textbf{95.73 $\pm$ 0.95} \\
\hline
\end{tabular}
} 
\end{table}

\begin{table}[H]
\centering
\caption{Classification results on SA training -- HU testing task}
\label{tab:HU}
\renewcommand{\arraystretch}{1.2}
\setlength{\tabcolsep}{6pt}
\resizebox{\textwidth}{!}{
\begin{tabular}{c|ccccccc}
\hline
\textbf{Metric} & \textbf{SSTN} & \textbf{CTF} & \textbf{DEMAE} & \textbf{DCFSL} & \textbf{FDFSL} & \textbf{HyMuT} & \textbf{Ours} \\
\hline
1 & 81.02 $\pm$ 5.04 & 89.62 $\pm$ 7.36 & 80.22 $\pm$ 10.45 & 90.22 $\pm$ 6.42 & 84.09 $\pm$ 5.36 & \textbf{90.27 $\pm$ 6.64} & 82.89 $\pm$ 10.23 \\
2 & 76.63 $\pm$ 13.89 & 74.81 $\pm$ 7.78 & 87.89 $\pm$ 6.56 & 81.44 $\pm$ 9.89 & 73.34 $\pm$ 12.52 & 81.41 $\pm$ 3.31 & \textbf{88.06 $\pm$ 5.07} \\
3 & \textbf{99.95 $\pm$ 0.07} & 98.70 $\pm$ 1.64 & 98.34 $\pm$ 1.78 & 94.77 $\pm$ 3.38 & 88.86 $\pm$ 7.86 & 96.24 $\pm$ 2.92 & 97.11 $\pm$ 3.25 \\
4 & \textbf{93.82 $\pm$ 0.43} & 90.93 $\pm$ 0.83 & 85.87 $\pm$ 6.85 & 90.02 $\pm$ 3.86 & 78.65 $\pm$ 4.15 & 90.14 $\pm$ 2.68 & 85.56 $\pm$ 4.83 \\
5 & 93.34 $\pm$ 7.12 & 93.86 $\pm$ 7.68 & 99.79 $\pm$ 0.48 & 94.86 $\pm$ 4.35 & 87.96 $\pm$ 6.38 & 98.67 $\pm$ 0.89 & \textbf{99.83 $\pm$ 0.39} \\
6 & 83.83 $\pm$ 3.01 & 86.56 $\pm$ 4.51 & \textbf{92.25 $\pm$ 6.23} & 76.25 $\pm$ 3.39 & 68.94 $\pm$ 4.84 & 74.81 $\pm$ 4.59 & 90.69 $\pm$ 3.69 \\
7 & 60.07 $\pm$ 12.22 & \textbf{79.47 $\pm$ 5.10} & 65.23 $\pm$ 15.28 & 72.46 $\pm$ 14.71 & 53.09 $\pm$ 9.74 & 67.25 $\pm$ 13.04 & 65.90 $\pm$ 10.38 \\
8 & 13.81 $\pm$ 8.55 & 44.15 $\pm$ 11.65 & \textbf{48.48 $\pm$ 8.03} & 34.50 $\pm$ 3.50 & 40.61 $\pm$ 9.64 & 40.31 $\pm$ 6.80 & 47.36 $\pm$ 8.60 \\
9 & 76.76 $\pm$ 12.22 & 77.17 $\pm$ 8.37 & 69.25 $\pm$ 9.14 & 60.45 $\pm$ 17.11 & 69.94 $\pm$ 7.42 & \textbf{77.32 $\pm$ 6.41} & 64.61 $\pm$ 9.17 \\
10 & 33.50 $\pm$ 5.00 & 60.97 $\pm$ 8.76 & 74.06 $\pm$ 16.75 & 60.03 $\pm$ 9.53 & 58.88 $\pm$ 12.28 & 55.73 $\pm$ 17.12 & \textbf{78.44 $\pm$ 13.88} \\
11 & 56.49 $\pm$ 24.74 & 57.07 $\pm$ 8.66 & \textbf{81.95 $\pm$ 13.58} & 58.18 $\pm$ 5.26 & 48.49 $\pm$ 8.29 & 61.11 $\pm$ 8.18 & 75.76 $\pm$ 12.10 \\
12 & 17.04 $\pm$ 4.97 & 70.85 $\pm$ 2.21 & 60.68 $\pm$ 10.07 & 70.54 $\pm$ 5.58 & 45.68 $\pm$ 8.93 & 63.99 $\pm$ 9.33 & \textbf{76.53 $\pm$ 12.40} \\
13 & 38.98 $\pm$ 32.77 & 92.17 $\pm$ 2.43 & 86.47 $\pm$ 8.90 & 90.99 $\pm$ 2.80 & 80.60 $\pm$ 8.75 & 87.11 $\pm$ 9.07 & \textbf{92.44 $\pm$ 3.80} \\
14 & 99.18 $\pm$ 0.99 & 91.88 $\pm$ 7.53 & \textbf{100.00 $\pm$ 0.00} & 91.63 $\pm$ 3.75 & 85.60 $\pm$ 8.10 & 88.09 $\pm$ 7.53 & 99.91 $\pm$ 0.22 \\
15 & \textbf{99.79 $\pm$ 0.29} & 98.37 $\pm$ 0.88 & 99.56 $\pm$ 0.87 & 92.95 $\pm$ 6.66 & 87.05 $\pm$ 5.37 & 95.15 $\pm$ 3.05 & 99.15 $\pm$ 2.03 \\
\hline
OA & 64.80 $\pm$ 4.23 & 77.50 $\pm$ 1.75 & 78.87 $\pm$ 3.70 & 74.63 $\pm$ 2.39 & 67.48 $\pm$ 2.51 & 75.67 $\pm$ 1.46 & \textbf{79.88 $\pm$ 1.75} \\
AA & 68.28 $\pm$ 3.83 & 80.44 $\pm$ 1.76 & 82.00 $\pm$ 3.18 & 77.29 $\pm$ 1.99 & 70.12 $\pm$ 2.37 & 77.84 $\pm$ 0.70 & \textbf{82.89 $\pm$ 1.35} \\
$\kappa\times100$ & 62.02 $\pm$ 4.67 & 75.69 $\pm$ 1.88 & 77.20 $\pm$ 3.99 & 72.58 $\pm$ 2.57 & 64.91 $\pm$ 2.70 & 73.71 $\pm$ 1.54 & \textbf{78.28 $\pm$ 1.89} \\
\hline
\end{tabular}
}
\end{table}

\begin{table}[H]
\centering
\caption{Classification results on HU training -- PU testing task}
\label{tab:PU}
\renewcommand{\arraystretch}{1.2}
\setlength{\tabcolsep}{6pt}
\resizebox{\textwidth}{!}{
\begin{tabular}{c|ccccccc}
\hline
\textbf{Metric} & \textbf{SSTN} & \textbf{CTF} & \textbf{DEMAE} & \textbf{DCFSL} & \textbf{FDFSL} & \textbf{HyMuT} & \textbf{Ours} \\
\hline
1 & 74.46 $\pm$ 8.53 & \textbf{83.20 $\pm$ 6.12} & 72.46 $\pm$ 8.64 & 89.57 $\pm$ 6.43 & 62.43 $\pm$ 11.08 & 77.64 $\pm$ 15.63 & 77.74 $\pm$ 9.10 \\
2 & 54.67 $\pm$ 18.14 & 65.05 $\pm$ 14.52 & 85.46 $\pm$ 7.93 & 78.77 $\pm$ 10.59 & 68.69 $\pm$ 7.75 & 78.71 $\pm$ 6.91 & \textbf{86.23 $\pm$ 7.02} \\
3 & 64.73 $\pm$ 29.69 & 67.19 $\pm$ 12.43 & 81.00 $\pm$ 11.10 & \textbf{98.15 $\pm$ 1.81} & 50.34 $\pm$ 12.98 & 61.45 $\pm$ 11.00 & 83.56 $\pm$ 7.68 \\
4 & 89.84 $\pm$ 3.03 & 87.88 $\pm$ 3.19 & 87.59 $\pm$ 3.84 & 88.56 $\pm$ 4.72 & 80.91 $\pm$ 8.48 & \textbf{90.83 $\pm$ 4.27} & 90.15 $\pm$ 3.79 \\
5 & 98.77 $\pm$ 1.74 & 99.65 $\pm$ 0.18 & 99.57 $\pm$ 0.59 & 96.41 $\pm$ 4.34 & 98.54 $\pm$ 0.98 & 96.09 $\pm$ 3.47 & \textbf{99.90 $\pm$ 0.21} \\
6 & 60.04 $\pm$ 31.67 & \textbf{91.67 $\pm$ 4.92} & 90.70 $\pm$ 6.80 & 79.50 $\pm$ 2.57 & 57.90 $\pm$ 8.15 & 66.48 $\pm$ 16.12 & 87.64 $\pm$ 6.79 \\
7 & 96.31 $\pm$ 0.94 & 96.91 $\pm$ 0.79 & 96.95 $\pm$ 2.00 & 74.90 $\pm$ 12.34 & 65.42 $\pm$ 12.08 & 79.14 $\pm$ 8.79 & \textbf{97.54 $\pm$ 1.63} \\
8 & 45.59 $\pm$ 35.66 & \textbf{78.73 $\pm$ 14.66} & 77.02 $\pm$ 9.25 & 41.29 $\pm$ 4.59 & 51.56 $\pm$ 11.13 & 65.29 $\pm$ 18.62 & 71.86 $\pm$ 13.47 \\
9 & 94.71 $\pm$ 3.61 & \textbf{99.01 $\pm$ 0.52} & 87.14 $\pm$ 9.26 & 70.28 $\pm$ 9.77 & 88.51 $\pm$ 7.81 & 99.41 $\pm$ 0.43 & 88.62 $\pm$ 9.22 \\
\hline
OA & 64.21 $\pm$ 3.18 & 76.73 $\pm$ 5.62 & 84.10 $\pm$ 2.76 & 76.72 $\pm$ 1.82 & 66.22 $\pm$ 3.52 & 76.99 $\pm$ 4.31 & \textbf{84.82 $\pm$ 2.94} \\
AA & 75.57 $\pm$ 4.89 & 84.81 $\pm$ 0.69 & 86.43 $\pm$ 2.27 & 79.25 $\pm$ 1.70 & 69.37 $\pm$ 2.75 & 79.45 $\pm$ 2.43 & \textbf{87.03 $\pm$ 1.61} \\
$\kappa\times100$ & 55.99 $\pm$ 3.14 & 71.14 $\pm$ 6.13 & 79.55 $\pm$ 3.23 & 74.83 $\pm$ 1.97 & 57.29 $\pm$ 3.81 & 70.40 $\pm$ 5.14 & \textbf{80.44 $\pm$ 3.51} \\
\hline
\end{tabular}
}
\end{table}

\begin{table}[H]
\centering
\caption{Classification results on PC training -- SA testing task}
\label{tab:SA}
\renewcommand{\arraystretch}{1.2}
\setlength{\tabcolsep}{6pt}
\resizebox{\textwidth}{!}{
\begin{tabular}{c|ccccccc}
\hline
\textbf{Metric} & \textbf{SSTN} & \textbf{CTF} & \textbf{DEMAE} & \textbf{DCFSL} & \textbf{FDFSL} & \textbf{HyMuT} & \textbf{Ours} \\
\hline
1 & 97.99 $\pm$ 1.42 & 99.63 $\pm$ 0.29 & 99.80 $\pm$ 0.55 & 98.28 $\pm$ 1.94 & 88.26 $\pm$ 4.44 & 94.71 $\pm$ 4.03 & \textbf{99.99 $\pm$ 0.03} \\
2 & 99.91 $\pm$ 1.42 & 99.96 $\pm$ 0.05 & \textbf{100.00 $\pm$ 0.00} & 95.62 $\pm$ 8.08 & 96.42 $\pm$ 1.99 & 98.84 $\pm$ 0.76 & 97.89 $\pm$ 4.09 \\
3 & 78.12 $\pm$ 17.85 & 97.24 $\pm$ 1.95 & \textbf{98.39 $\pm$ 4.69} & 98.04 $\pm$ 1.56 & 82.52 $\pm$ 10.01 & 91.56 $\pm$ 7.20 & 97.81 $\pm$ 3.08 \\
4 & 98.01 $\pm$ 0.92 & 99.21 $\pm$ 0.60 & 98.76 $\pm$ 1.21 & 98.76 $\pm$ 1.33 & 96.98 $\pm$ 1.69 & 99.06 $\pm$ 0.93 & \textbf{99.47 $\pm$ 0.79} \\
5 & \textbf{97.05 $\pm$ 1.26} & 93.83 $\pm$ 1.80 & 92.26 $\pm$ 5.15 & 92.01 $\pm$ 0.92 & 81.15 $\pm$ 6.36 & 91.16 $\pm$ 3.33 & 93.32 $\pm$ 4.07 \\
6 & 97.86 $\pm$ 1.29 & \textbf{99.94 $\pm$ 0.04} & 98.28 $\pm$ 2.01 & 99.38 $\pm$ 0.73 & 97.73 $\pm$ 1.33 & 97.92 $\pm$ 2.11 & 99.55 $\pm$ 0.84 \\
7 & 99.81 $\pm$ 0.06 & 99.85 $\pm$ 0.13 & \textbf{99.93 $\pm$ 0.13} & 99.50 $\pm$ 0.21 & 98.19 $\pm$ 0.93 & 99.17 $\pm$ 0.70 & \textbf{99.93 $\pm$ 0.12} \\
8 & 53.78 $\pm$ 10.48 & 80.89 $\pm$ 3.53 & 78.00 $\pm$ 6.63 & 67.27 $\pm$ 10.68 & 69.50 $\pm$ 7.57 & 71.88 $\pm$ 8.63 & \textbf{82.26 $\pm$ 8.13} \\
9 & 98.27 $\pm$ 1.07 & 99.63 $\pm$ 0.45 & \textbf{100.00 $\pm$ 0.00} & 98.93 $\pm$ 1.39 & 96.74 $\pm$ 1.63 & 98.56 $\pm$ 1.45 & 99.95 $\pm$ 0.11 \\
10 & 86.37 $\pm$ 10.48 & 87.14 $\pm$ 1.25 & 92.24 $\pm$ 3.17 & 84.98 $\pm$ 4.39 & 70.57 $\pm$ 11.63 & 82.49 $\pm$ 10.40 & \textbf{95.45 $\pm$ 3.14} \\
11 & 90.71 $\pm$ 1.72 & 97.37 $\pm$ 2.39 & \textbf{99.97 $\pm$ 0.08} & 98.34 $\pm$ 1.06 & 89.45 $\pm$ 3.71 & 96.41 $\pm$ 1.98 & \textbf{99.97 $\pm$ 0.06} \\
12 & \textbf{99.81 $\pm$ 0.24} & 99.71 $\pm$ 0.09 & 97.86 $\pm$ 2.77 & 99.35 $\pm$ 0.86 & 99.01 $\pm$ 1.01 & 99.65 $\pm$ 0.40 & 98.30 $\pm$ 2.54 \\
13 & \textbf{99.74 $\pm$ 0.19} & 99.09 $\pm$ 0.72 & 99.43 $\pm$ 0.63 & 99.43 $\pm$ 0.55 & 98.19 $\pm$ 1.50 & 99.32 $\pm$ 0.32 & 99.62 $\pm$ 0.65 \\
14 & 93.21 $\pm$ 2.46 & \textbf{99.00 $\pm$ 0.77} & 96.50 $\pm$ 4.99 & 98.42 $\pm$ 1.09 & 97.95 $\pm$ 1.30 & 97.71 $\pm$ 0.83 & 93.15 $\pm$ 7.69 \\
15 & \textbf{87.21 $\pm$ 7.96} & 78.15 $\pm$ 15.30 & 86.43 $\pm$ 8.16 & 67.47 $\pm$ 6.60 & 70.81 $\pm$ 5.76 & 73.59 $\pm$ 5.20 & 84.48 $\pm$ 13.01 \\
16 & 97.97 $\pm$ 1.09 & 91.05 $\pm$ 5.66 & \textbf{97.99 $\pm$ 3.35} & 91.79 $\pm$ 6.53 & 82.37 $\pm$ 6.31 & 84.71 $\pm$ 7.60 & 97.83 $\pm$ 3.58 \\
\hline
OA & 85.96 $\pm$ 1.90 & 91.41 $\pm$ 1.59 & 92.29 $\pm$ 0.93 & 86.46 $\pm$ 2.30 & 84.05 $\pm$ 1.51 & 87.46 $\pm$ 1.25 & \textbf{93.05 $\pm$ 0.65} \\
AA & 92.24 $\pm$ 1.27 & 95.10 $\pm$ 0.87 & 95.99 $\pm$ 0.80 & 92.97 $\pm$ 0.42 & 88.49 $\pm$ 1.02 & 92.30 $\pm$ 1.11 & \textbf{96.19 $\pm$ 0.68} \\
$\kappa\times100$ & 84.47 $\pm$ 2.06 & 90.44 $\pm$ 1.78 & 91.44 $\pm$ 1.03 & 84.97 $\pm$ 2.52 & 82.30 $\pm$ 1.64 & 86.07 $\pm$ 1.37 & \textbf{92.28 $\pm$ 0.72} \\
\hline
\end{tabular}
}
\end{table}

\subsection{Parameter Analysis}

A systematic parameter analysis is performed, where comparative evaluation of classification performance across diverse parameter settings in cross-domain scenarios enables performance-based optimization.

1) Band masking ratio (\textit{masking\_ratio})

The masking ratio critically affects spectral feature learning. Insufficient masking (ratio\(=\)0.1) weakens constraint effectiveness, whereas excessive masking (ratio\(=\)0.5) fractures spectral continuity. Quantitative evidence shows ratio elevation from 0.1 to 0.3 boosts SA\(\rightarrow\)HU performance (OA +2.54\%, AA +2.04\%, and Kappa +2.74\%), but further raising it to 0.5 degrades HU\(\rightarrow\)PU metrics (OA -2.50\%, AA -1.72\%, Kappa -0.16\%). 

2) Number of bidirection cross-attention layers (\textit{Bicrossattention\_num})

It directly determines the interaction capacity between spatial and spectral features in S$^2$Former, which exhibits a non-monotonic relationship with model performance. The two-layer configuration represents the computational sweet spot, with 1\(\rightarrow\)2 layer increase boosting HU\(\rightarrow\)PU metrics by 0.99\% OA, 1.15\% AA and 1.28\% Kappa, while 2\(\rightarrow\)3 layer increase causes 3.25\% OA, 2.98\% AA and 4.08\% Kappa degradation. Single-layer interactions inadequately activate complementary features, whereas three-layer design induces overparameterization. The two-layer architecture optimally balances classification accuracy and stability.

3) Training Sample Size

The sample size scalability experiment examines cross-domain performance under varying target domain quantities (from 5 to 25 samples per class with 5-sample intervals). Under consistent experimental conditions, our method demonstrates progressive performance gains with increasing sample size across all transfer tasks, surpassing all comparative methods in sample efficiency. 

\begin{table}[ht]
\centering
\caption{Performance comparison of different parameter settings}
\resizebox{\textwidth}{!}{
\begin{tabular}{ccc|ccc|cccc}
\hline
\multicolumn{3}{c|}{} & \multicolumn{3}{c|}{\textbf{\textit{Masking\_ratio}}} & \multicolumn{3}{c}{\textbf{\textit{Bicrossattention\_num}}} \\
\hline
\textbf{Source} & \textbf{Target} & \textbf{Metric} & 0.1 & Ours (0.3) & 0.5 & 1 & Ours (2) & 3 \\
\hline
&   & OA & 96.18 $\pm$ 0.85 & \textbf{96.98 $\pm$ 0.67} & 96.63 $\pm$ 0.61 & 96.71 $\pm$ 1.03 & \textbf{96.98 $\pm$ 0.67} & 96.35 $\pm$ 0.99 \\
PU & PC & AA & 91.87 $\pm$ 1.77 & \textbf{93.30 $\pm$ 1.34} & 92.73 $\pm$ 1.50 & 92.98 $\pm$ 1.87 & \textbf{93.30 $\pm$ 1.34} & 92.75 $\pm$ 1.53 \\
& & $\kappa\times100$ & 94.61 $\pm$ 1.20 & \textbf{95.73 $\pm$ 0.95} & 95.24 $\pm$ 0.86 & 95.35 $\pm$ 1.44 & \textbf{95.73 $\pm$ 0.95} & 94.85 $\pm$ 1.39 \\
\hline
&    & OA & 77.34 $\pm$ 2.85 & \textbf{79.88 $\pm$ 1.75} & 78.99 $\pm$ 3.10 & 78.18 $\pm$ 1.55 & \textbf{79.88 $\pm$ 1.75} & 78.29 $\pm$ 3.83 \\
SA & HU & AA & 80.85 $\pm$ 2.43 & \textbf{82.89 $\pm$ 1.35} & 82.23 $\pm$ 2.63 & 81.37 $\pm$ 1.30 & \textbf{82.89 $\pm$ 1.35} & 81.55 $\pm$ 3.20 \\
& & $\kappa\times100$ & 75.54 $\pm$ 3.07 & \textbf{78.28 $\pm$ 1.89} & 77.32 $\pm$ 3.34 & 76.44 $\pm$ 1.67 & \textbf{78.28 $\pm$ 1.89} & 76.56 $\pm$ 4.14 \\
\hline
&  & OA & 83.36 $\pm$ 3.51 & \textbf{84.82 $\pm$ 2.94} & 82.32 $\pm$ 3.39 & 83.83 $\pm$ 4.23 & \textbf{84.82 $\pm$ 2.94} & 81.57 $\pm$ 3.22 \\
HU & PU & AA & 86.33 $\pm$ 1.40 & \textbf{87.03 $\pm$ 1.61} & 85.31 $\pm$ 2.19 & 85.88 $\pm$ 3.10 & \textbf{87.03 $\pm$ 1.61} & 84.05 $\pm$ 2.95 \\
& & $\kappa\times100$ & 78.65 $\pm$ 4.07 & \textbf{80.44 $\pm$ 3.51} & 77.28 $\pm$ 3.91 & 79.16 $\pm$ 5.09 & \textbf{80.44 $\pm$ 3.51} & 76.36 $\pm$ 3.85 \\
\hline
 &  & OA & 92.93 $\pm$ 0.83 & \textbf{93.05 $\pm$ 0.65} & 92.54 $\pm$ 1.22 & 92.46 $\pm$ 1.43 & \textbf{93.05 $\pm$ 0.65} & 92.72 $\pm$ 1.03 \\
PC & SA & AA & 95.73 $\pm$ 0.99 & \textbf{96.19 $\pm$ 0.68} & 95.76 $\pm$ 0.99 & 95.84 $\pm$ 0.88 & \textbf{96.19 $\pm$ 0.68} & 95.94 $\pm$ 0.89 \\
& & $\kappa\times100$ & 92.15 $\pm$ 0.91 & \textbf{92.28 $\pm$ 0.72} & 91.72 $\pm$ 1.35 & 91.65 $\pm$ 1.58 & \textbf{92.28 $\pm$ 0.72} & 91.91 $\pm$ 1.13 \\
\hline
\end{tabular}
}
\end{table}

\begin{figure}[H]
    \centering
    \captionsetup{belowskip=0pt}
    \includegraphics[width=\textwidth]{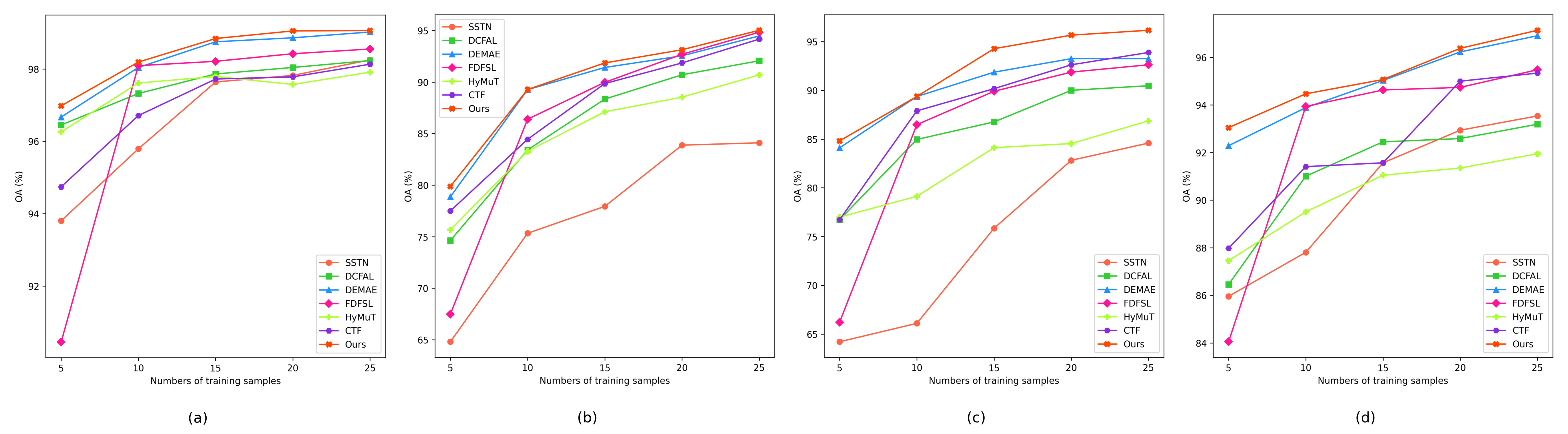}
    \caption{OA values for different sample sizes and models}
    \label{fig:OA_values}
\end{figure}

\subsection{Ablation experiments and visualization}

Ablation experiments assessing module contributions sequentially remove S²Former, FDC, and DAFT module, comparing variants with the complete model across four tasks. The t-SNE visualizations reveal the feature clustering effects after data preprocessing and classification.

1) Ablating S²Former causes in the most severe performance deterioration, with HU\(\rightarrow\)PU metrics dropping by 3.34\% OA and 4.29\% Kappa, with similar trends observed across other tasks. This confirms its indispensable spatial--spectral modeling capability. The module's bidirectional attention mechanism enables dynamic fusion of spatial and spectral information, capturing both local patterns and global dependencies. Its removal cripples cross-modal integration, impairing feature alignment and classification accuracy.

2) FDC ablation demonstrates context-dependent effectiveness: it degrades performance in spectrally complex tasks (SA\(\rightarrow\)HU: -1.14\% OA, -1.23\% Kappa, similar trends observed in PU\(\rightarrow\)PC and PC\(\rightarrow\)SA tasks) while slightly improving simpler scenarios (HU\(\rightarrow\)PU: +0.34\% OA, +0.22\% Kappa). This contrast confirms FDC's specialization in spectral detail preservation and high-frequency modeling for complex targets. The performance gain in HU\(\rightarrow\)PU suggests feature smoothing benefits in structurally simple environments, revealing the module's adaptive value based on spectral-structural complexity.

3) DAFT proves critical for domain alignment, with its removal causing severe performance drop in the HU\(\rightarrow\)PU task (6.66\% OA, 7.77\% Kappa). The module enables cross-domain feature calibration through collaborative alignment, preventing semantic drift and maintaining distribution balance. Its absence disrupts transfer stability, confirming its role in domain adaption.

\begin{figure}[H]
    \centering
    \captionsetup{belowskip=0pt}
    \includegraphics[width=\textwidth]{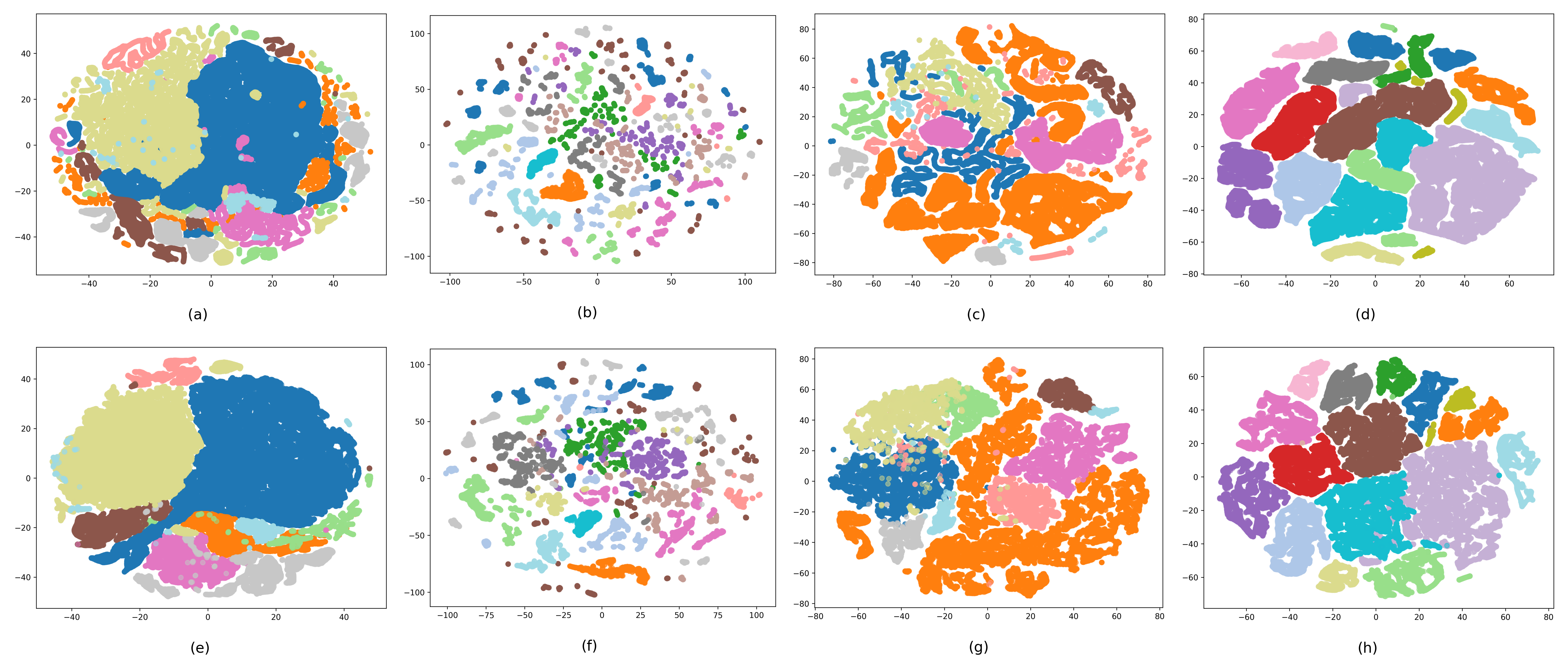}
    \caption{t-SNE visualization of the preprocessing and classification features on four target domain datasets.(a) PC dataset after preprocessing, (b) HU dataset after preprocessing, (c) PU dataset after preprocessing, (d) SA dataset after preprocessing. (e) PC dataset after classification, (f) HU dataset after classification, (g) PU dataset after classification, (h) SA dataset after classification.}
    \label{fig:t-SNE}
\end{figure}

4) As shown in Figure~\ref{fig:t-SNE}, our model produces tightly clustered features and distinct boundaries in the target domain, successfully separating spectrally similar categories (e.g., categories 7 and 8 in PC, categories 2 and 5 in HU) and refining ambiguous categories (e.g., categories 8 and 10 in SA, categories 2 and 7 in PU) through improved intra-class clustering and inter-class separation after self-supervised transfer, demonstrating superior discriminative capability.

\begin{table}[H]
\centering
\caption{Ablation study results between different modules}
\label{tab:melting}
\renewcommand{\arraystretch}{1}
\setlength{\tabcolsep}{6pt}
\resizebox{\textwidth}{!}{
\begin{tabular}{cccccccc}
\hline
\textbf{Source} & \textbf{Target} & \textbf{Metric} & \textbf{w/o S²Former} & \textbf{w/o FDC} & \textbf{w/o DAFT} & \textbf{Ours} \\
\hline
   &    & OA & 96.01 $\pm$ 1.10 & 96.29 $\pm$ 1.21 & 96.32 $\pm$ 1.20 & \textbf{96.98 $\pm$ 0.67} \\
PU & PC & AA & 91.56 $\pm$ 1.21 & 92.07 $\pm$ 2.38 & 92.49 $\pm$ 1.46 & \textbf{93.30 $\pm$ 1.34} \\
   &    & $\kappa\times100$ & 94.37 $\pm$ 1.53 & 94.77 $\pm$ 1.68 & 94.82 $\pm$ 1.65 & \textbf{95.73 $\pm$ 0.95} \\
\hline
   &    & OA & 77.20 $\pm$ 2.21 & 78.74 $\pm$ 1.24 & 78.06 $\pm$ 2.99 & \textbf{79.88 $\pm$ 1.75} \\
SA & HU & AA & 80.53 $\pm$ 1.76 & 81.81 $\pm$ 1.08 & 81.26 $\pm$ 2.56 & \textbf{82.89 $\pm$ 1.35} \\
   &    & $\kappa\times100$ & 75.38 $\pm$ 2.29 & 77.05 $\pm$ 1.33 & 76.31 $\pm$ 3.23 & \textbf{78.28 $\pm$ 1.89} \\
\hline
   &    & OA & 81.48 $\pm$ 1.59 & \textbf{85.16 $\pm$ 2.36} & 78.16 $\pm$ 5.15 & 84.82 $\pm$ 2.94 \\
HU & PU & AA & 81.70 $\pm$ 1.16 & 85.86 $\pm$ 2.40 & 84.73 $\pm$ 2.04 & \textbf{87.03 $\pm$ 1.61} \\
   &    & $\kappa\times100$ & 76.15 $\pm$ 1.94 & \textbf{80.66 $\pm$ 3.03} & 72.67 $\pm$ 5.74 & 80.44 $\pm$ 3.51 \\
\hline
   &    & OA & 92.26 $\pm$ 0.72 & 92.79 $\pm$ 0.31 & 92.17 $\pm$ 0.63 & \textbf{93.05 $\pm$ 0.65} \\
PC & SA & AA & 96.02 $\pm$ 0.77 & 95.87 $\pm$ 0.81 & 95.44 $\pm$ 1.00 & \textbf{96.19 $\pm$ 0.68} \\
   &    & $\kappa\times100$ & 91.41 $\pm$ 0.80 & 91.99 $\pm$ 0.34 & 91.30 $\pm$ 0.70 & \textbf{92.28 $\pm$ 0.72} \\
\hline
\end{tabular}
}
\end{table}

\section{Conclusion}\label{conclusion}
A novel self-supervised cross-domain transfer framework is proposed in this paper, which successfully addresses the challenge of cross-domain transfer in hyperspectral image classification by learning transferable spectral-spatial joint representations without label supervision. For self-supervised pre-training, the designed S²Former module incorporates dual-branch transformer operating on spatial and spectral domains, enhanced by bidirectional cross-attention for effective spectral-spatial cooperative modeling. The spatial branch enhances structural awareness through random masking, while the spectral branch captures fine-grained differences. These branches mutually guide each other to improve semantic consistency. The FDC strengthens the model's capacity to discern details and boundaries by preserving frequency-domain structural consistency. During fine-tuning, the proposed DAFT mechanism effectively aligns semantic evolution trajectories, enabling robust transfer learning under low-label conditions.

Experimental results indicate stable classification performance and excellent cross-domain adaptability across four hyperspectral datasets, confirming its efficacy under resource-constrained conditions. Nevertheless, the current approach focuses solely on pre-existing semantic features in the target domain. Future research could explore specific characteristics of unlabeled data within the target domain through self-supervised methods like contrastive learning to substantially improve cross-domain transfer resilience and adaptive performance.


\bibliography{mybibfile}

\end{document}